\documentclass[10pt,twocolumn]{article} 
\usepackage{preprint}
\usepackage{times}
\usepackage{graphicx}
\usepackage{amssymb}
\usepackage{amsmath}
\usepackage{amsthm}
\usepackage{url}
\usepackage[table]{xcolor}
\usepackage{enumitem}  
\usepackage{natbib}
\usepackage{tabularx}
\usepackage{booktabs}
\usepackage[hidelinks]{hyperref}
\usepackage{amsmath}
\usepackage{tikz}
\usetikzlibrary{arrows.meta, positioning, calc}
\usepackage{amsthm}
\usepackage{fancyhdr}
\usepackage[hidelinks]{hyperref}
\usepackage{url}
\usepackage[english]{babel}

\newcommand{\githubalgorithm}{https://github.com/INFORMSJoC/2023.0419}
\newcommand{\githubresults}{https://github.com/INFORMSJoC/2023.0419}
\newcommand{\githubdata}{https://github.com/INFORMSJoC/2023.0419}

\newcommand{\vect}[1]{\boldsymbol{#1}}
\newtheorem{theorem}{Theorem}
\newtheorem{lemma}[theorem]{Lemma}

\begin{document}

\title{An algorithm for clustering with confidence-based must-link and cannot-link constraints}

\author{P.~Baumann$^1$, D.~S.~Hochbaum$^2$\\
	\\
	$^1$Department of Business Administration, University of Bern, Engehaldenstr.\ 4, 3012 Bern, Switzerland\\
	$^2$IEOR Department, University of California, Berkeley, CA 94720, USA \\	
	\\
	philipp.baumann@unibe.ch, dhochbaum@berkeley.edu
	\\
	\\
	This work has been published in the \href{https://pubsonline.informs.org/doi/10.1287/ijoc.2023.0419}{\color{blue}INFORMS Journal on Computing}
	\\
	\url{https://doi.org/10.1287/ijoc.2023.0419}
	\\
	The code of the algorithm is being developed on an on-going basis at \href{https://github.com/phil85/PCCC-Algorithm}{\color{blue}github.com/phil85/PCCC-Algorithm}.
	}

\maketitle
\thispagestyle{empty}

% Configure the fancyhdr settings
\fancypagestyle{empty}{
	\fancyhf{} % Clear all headers and footers
	\renewcommand{\headrulewidth}{0pt} % Remove header rule
	\renewcommand{\footrulewidth}{0pt} % Remove footer rule
	\fancyfoot[C]{\copyright\ 2024 D.S.\ Hochbaum, P.\ Baumann, O.\ Goldschmidt and Y.\ Zhang.} % Centered footer
}

\bibliographystyle{elsarticle-harv}

\begin{abstract}
We study here the semi-supervised $k$-clustering problem where information is available on whether pairs of objects are in the same or in different clusters. This information is either available with certainty or with a limited level of confidence. We introduce the PCCC (Pairwise-Confidence-Constraints-Clustering) algorithm, which iteratively assigns objects to clusters while accounting for the information provided on the pairs of objects. Our algorithm uses integer programming for the assignment of objects which allows to include relationships as hard constraints that are guaranteed to be satisfied or as soft constraints that can be violated subject to a penalty. This flexibility distinguishes our algorithm from the state-of-the-art in which all pairwise constraints are either considered hard, or all are considered soft. We developed an enhanced multi-start approach and a model-size reduction technique for the integer program that contributes to the effectiveness and the efficiency of the algorithm. Unlike existing algorithms, our algorithm scales to large-scale instances with up to 60,000 objects, 100 clusters, and millions of cannot-link constraints (which are the most challenging constraints to incorporate). We compare the PCCC algorithm with state-of-the-art approaches in an extensive computational study. Even though the PCCC algorithm is more general than the state-of-the-art approaches in its applicability, it outperforms the state-of-the-art approaches on instances with all hard or all soft constraints both in terms of runtime and various metrics of solution quality. The code of the PCCC algorithm is publicly available on \href{\githubalgorithm}{\color{blue}GitHub}.
\end{abstract}

\section{Introduction}
\noindent Clustering is a fundamental task in knowledge discovery with diverse applications, including data exploration, anomaly detection, and feature engineering. Clustering algorithms have been found to deliver better results when provided with additional information (see \citealt{valls2009using}). We consider here a semi-supervised $k$-clustering problem where additional information is available in the form of must-link and cannot-link constraints, indicating pairs of objects belonging to the same or different clusters, respectively. These pairwise relationships are either available with certainty (hard constraints) or with a limited level of confidence (soft constraints). There is a wide range of clustering applications with must-link and cannot-link constraints, such as facility location, genomics, image segmentation, and text analytics (see, e.g., \citealt{pelegrin2023new, tian2021model, zhang2022semi, yang2022analyzing}). One real-world clustering application that involves hard and soft must-link and cannot-link constraints stems from astronomical image analysis. In this application, experts provide information with different levels of confidence on whether objects in different telescope images correspond to the same celestial body or different ones. 

Various algorithms have been proposed for clustering with hard or soft must-link and cannot-link constraints (e.g.,~\citealt{chen2014clustering, ganji2016lagrangian, le2019binary, gonzalez2020dils, piccialli2022exact}). Most of these algorithms can be divided into three groups. The first group comprises exact approaches that employ constraint programming, column generation, semidefinite and integer programming techniques. Exact approaches have the advantage that they are guaranteed to find an assignment that satisfies all hard constraints (if such a feasible assignment exists), but due to their prohibitive computational cost, they can solve only small instances with up to 1,000 objects and a small number of clusters within reasonable running time. The second group comprises \emph{center-based} heuristics that represent a cluster by a single representative (often the center of gravity or the median of the cluster). Center-based heuristics assign objects sequentially in random order while considering the pairwise constraints. These heuristics are generally fast, but the sequential assignment strategies often fail to find high-quality or even feasible solutions when the number of pairwise constraints (in particular, the number of cannot-link constraints) is large. The third group comprises metaheuristics that improve a solution by modifying an assignment vector with different randomized operators. In the presence of many pairwise constraints, randomized modifications of the assignment vectors often increase the total number of constraint violations making it challenging to find improvements. Most existing algorithms that treat pairwise relationships as soft constraints only treat them as soft constraints to attain feasible solutions and deal with the computational difficulty of hard constraints. Therefore, these algorithms assign very high penalties for violating soft constraints; hence, the soft constraints are only violated if these constraints cannot be satisfied. Consequently, these algorithms are not suitable for instances where the confidence of some relationships is limited. 

We propose the PCCC \emph{Pairwise-Confidence-Constraints-Clustering} algorithm, which is a center-based heuristic that alternates between an object assignment and a cluster center update step like the k-means algorithm. The key idea that separates our algorithm from existing center-based heuristics is that it uses integer programming instead of a sequential assignment procedure to accomplish the object assignment step. Using integer programming for the entire problem is intractable, even for small instances, but when used to solve the assignment step only, with our specific enhancements, it scales to very large instances. A mixed binary linear optimization problem (MBLP) is solved in each assignment step. The MBLP is formulated such that hard and soft constraints can be considered simultaneously. To reduce the size of the MBLP, we apply a preprocessing procedure that contracts objects linked by hard must-link constraints. The MBLP accounts for constraint-specific weights that reflect the degree of confidence in soft constraints. Additional constraints, such as cardinality or fairness constraints, can be easily incorporated into the MBLP, which makes the PCCC algorithm applicable to a wide range of constrained clustering problems with minor modifications. 

An important parameter that controls the computational efficiency of our algorithm is to permit assignments of objects only to the $q \ll k$ closest cluster centers. Instead of applying this parameter uniformly to all objects, we identify during the process a subset of objects which is ``critical", meaning that the objects in the subset cause the largest penalties related to violations of soft cannot-link constraints. For these critical objects we adjust upwards the value of $q$. This approach enables generating better solutions while still maintaining low computational effort. 

Furthermore, we introduce a new concept for multi-start methods that greatly improves the effectiveness of our method. Rather than terminating an iteration after convergence, we apply specific modifications to the converged solution and use this modified solution to continue the iteration. For constrained clustering this modification involves repositioning of some of the cluster centers.

We present a computational study on the effectiveness of the PCCC algorithm in terms of solution quality and running time. This study compares the PCCC algorithm to the exact approach of \cite{piccialli2022exact} and four state-of-the-art heuristic algorithms. We use several performance measures to evaluate the solution quality. The test set includes benchmark instances from the literature, synthetic instances that allow studying the impact of complexity parameters, and large-scale instances that allow to assess the scalability of the algorithms. In addition, we investigate the benefits of specifying weights for soft constraints to express different confidence levels. The experiments demonstrate that the PCCC algorithm is able to find optimal or near-optimal solutions for small instances within few seconds and that it outperforms the state-of-the-art algorithms in terms of ARI values, Inertia (within cluster sum-of-squares), Silhouette coefficients, and running time. Furthermore, the use of constraint-specific weights in the PCCC algorithm is shown to be particularly beneficial when an instance comprises a large number of noisy (incorrect) pairwise constraints.

The paper is structured as follows. In Section~\ref{sec_problem}, we provide a formal description and an illustrative example of the constrained clustering problem studied here. In Section~\ref{sec_literature}, we discuss the related literature. In Section~\ref{sec_algorithm}, we explain the components of the PCCC algorithm and introduce the algorithmic enhancements. In Sections~\ref{sec_experiment} and \ref{sec_experiments_comparison_noisy}, we report the results of the computational comparison and analyze the benefits of accounting for confidence values, respectively. Finally, in Section~\ref{sec_conclusions}, we summarize the paper and provide ideas for future research. 
	
\section{Constrained clustering}\label{sec_problem}
Different variants of constrained clustering problems have been studied in the literature. In Section~\ref{sec_problem_description}, we describe our problem that generalizes problem variants considered in the literature. In Section~\ref{sec_problem_example}, we provide the data of an illustrative example that we will use later in the paper to illustrate the PCCC algorithm.

\subsection{The constrained clustering problem}\label{sec_problem_description}
Consider a data set with $n$ objects $\vect{x}_1, \ldots, \vect{x}_n \in \mathbb{R}^d$, where each object is a vector in a $d$-dimensional Euclidean feature space, and an integer $k$ designating the number of clusters. The goal is to assign a label $l_i \in \{1, 2, \ldots, k\}$ to each object $i=1, 2, \ldots,n$ such that some clustering objective is optimized. Additional information is given in the form of pairwise must-link and cannot-link constraints, which are either hard constraints that must be satisfied in a feasible assignment or soft constraints that may be violated. We denote the set of pairs of objects that are subject to a hard must-link constraint as $ML$, and the set of pairs of objects that are subject to a hard cannot-link constraint as $CL$. The set of pairs of objects that are subject to soft must-link constraints are denoted by $SML$, and the set of pairs of objects that are subject to a soft cannot-link constraint are denoted by $SCL$. Each pair $\{i, j\} \in SML~\cup~SCL$ is associated with a confidence value $w_{ij} \in (0, 1]$, which represents the degree of belief in the corresponding information. The higher $w_{ij}$, the higher the degree of belief. Note that the pairs are unordered such that $\{i, j\}=\{j, i\}$. Since different algorithms may use different clustering objectives, we use three generally accepted metrics to evaluate clustering quality. The Inertia criterion (or sum-of-squares criterion) which measures how internally coherent the clusters are, the Silhouette coefficient of \cite{rousseeuw1987silhouettes}, which quantifies both the intra-cluster cohesion and inter-cluster separation, and the Adjusted Rand Index (ARI) of \cite{hubert1985comparing} which compares the assigned labels $l_i$ with the ground truth labels $y_i$ while ignoring permutations and adjusting for chance.

\subsection{Illustrative example}\label{sec_problem_example}
\begin{figure*}
	\centering\includegraphics[width=0.45\textwidth]{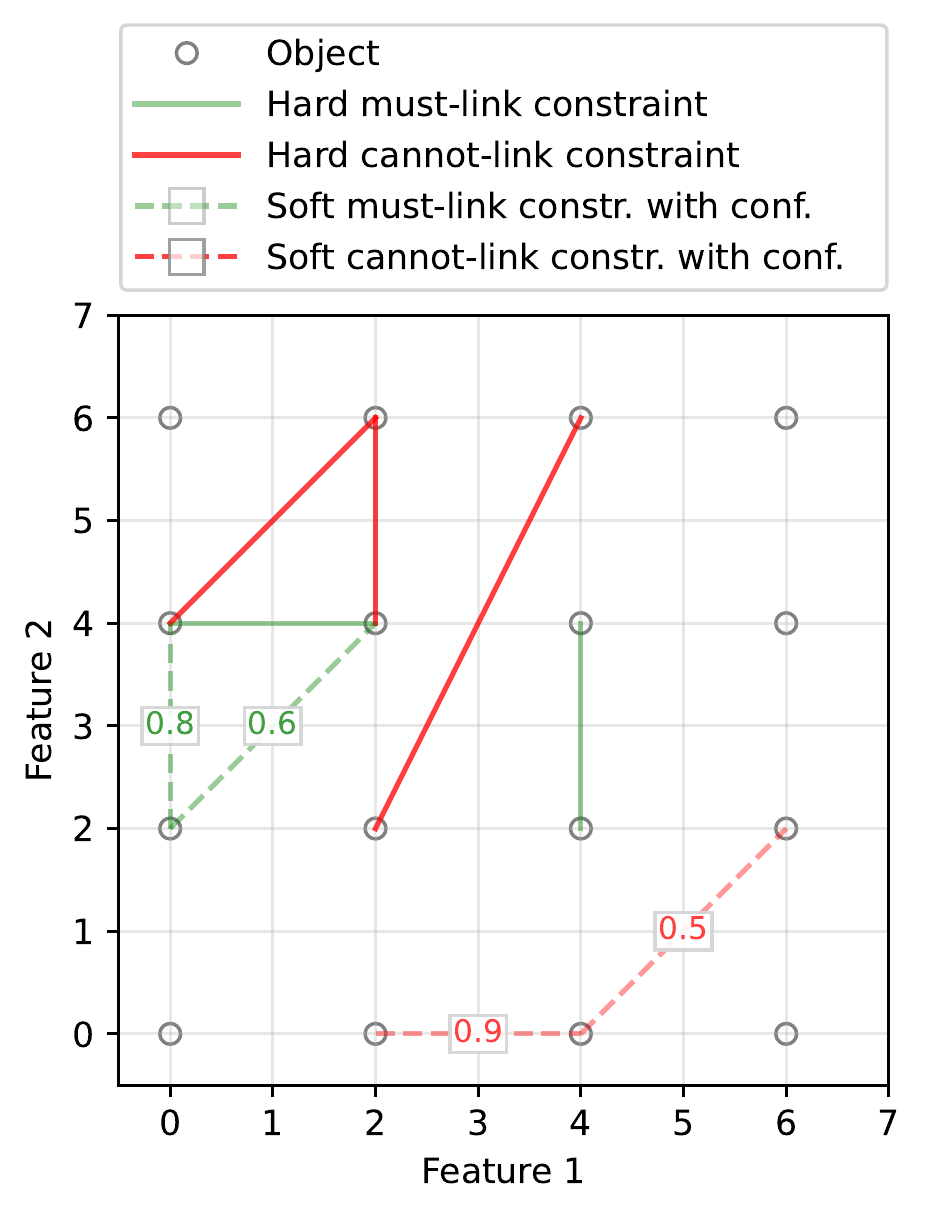}\includegraphics[width=0.45\textwidth]{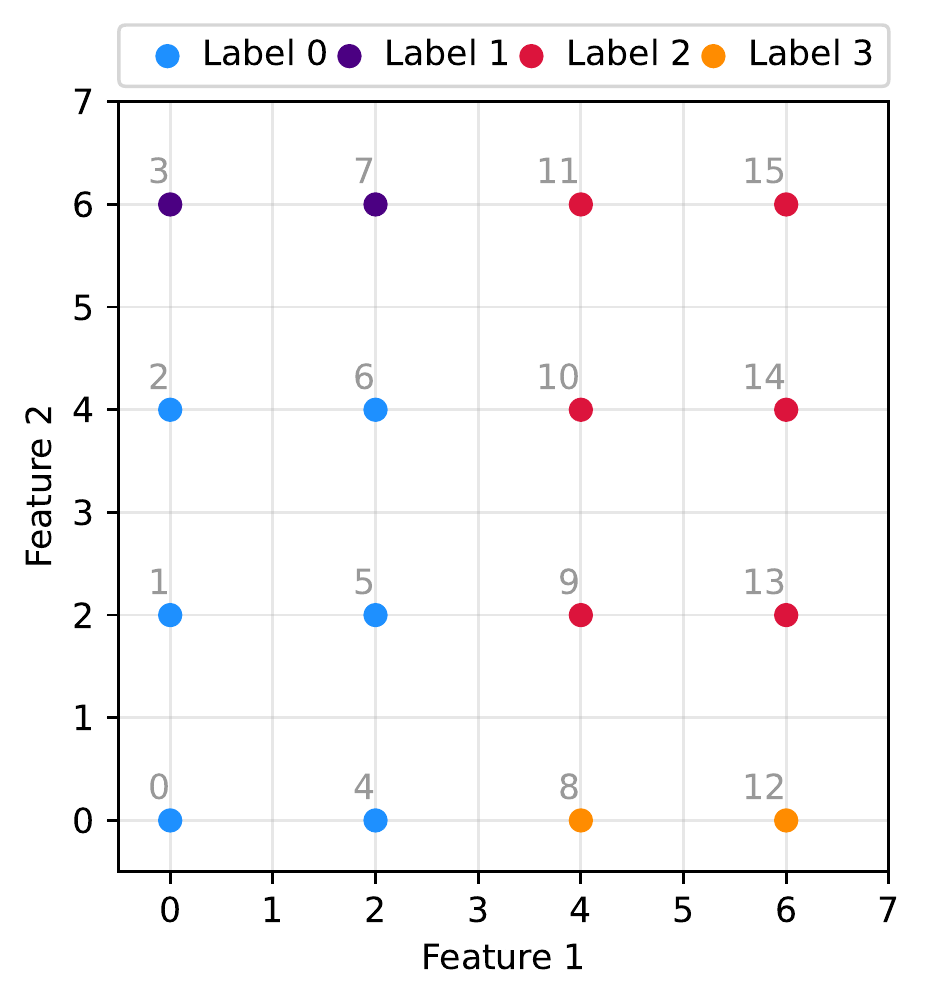}
	\caption{Illustrative example: input data (left) and ground truth (right).\label{fig_ie_data}}
\end{figure*}

We consider a data set with $n=16$ objects that are described by $d=2$ features. The goal is to partition this data set into $k=4$ clusters. The subplot on the left-hand side of Figure~\ref{fig_ie_data} visualizes the data set and the additional information which consists of three hard cannot-link constraints, two hard must-link constraints, two soft cannot-link constraints, two soft must-link constraints, and also the respective confidence levels. The ground truth assignment is shown in the subplot on the right-hand side of Figure~\ref{fig_ie_data}. 

\section{Related literature}\label{sec_literature}
Many different algorithms have been proposed for clustering with must-link and cannot-link constraints. We focus our review on clustering algorithms that take the number of clusters $k$ as an input parameter and assign each object to exactly one cluster. We categorize the algorithms according to their methodology into exact, center-based, and non-center based algorithms. The three groups are discussed in Sections~\ref{sec_literature_exact_approaches}--\ref{sec_literature_other}. Table~\ref{tab_literature} lists for each algorithm the methodology used, the clustering objective (e.g., k-means), and whether the objects are assigned sequentially to clusters or simultaneously. The checkmarks ''$\checkmark$" in the table indicate which features of the constrained clustering problem described in the previous section are covered by the respective algorithms. A checkmark enclosed in paranthesis ''($\checkmark$)" indicates that the algorithm treats the respective constraints as hard constraints, but might still return a clustering where some of these constraints are violated. A checkmark in the column ``user-defined penalty'' means the user can specify a single penalty value for constraint violations. Algorithms that allow the user to define penalty values individually for the soft must-link and cannot-link constraint have an additional checkmark in the column ``pairwise-specific penalties''. The approaches from the literature that we test in our experiments are highlighted in bold, both in the text and in Table~\ref{tab_literature}.

\begin{table*}
	\footnotesize
	\setlength{\tabcolsep}{4pt}
	\begin{tabularx}{\textwidth}{Xlllcccccc}
		\toprule
		\textbf{Name}  & \textbf{Method} & \textbf{Objective} & \textbf{Object}  & \textbf{Hard}  & \textbf{Hard}  & \textbf{Soft} & \textbf{Soft}  & \textbf{User-}   & \textbf{Pairwise-} \\
		\textbf{}      & \textbf{}       & \textbf{}          & \textbf{assign-} & \textbf{ML}    & \textbf{CL}    & \textbf{ML}   & \textbf{CL}    & \textbf{defined} & \textbf{specific}  \\
		\textbf{}      & \textbf{}       & \textbf{}          & \textbf{ment}    & \textbf{}      & \textbf{}      & \textbf{}     & \textbf{}      & \textbf{penalty} & \textbf{penalties} \\ \midrule
		CP             & exact           & multiple           & simult.          & $\checkmark$   & $\checkmark$   &               &                &                  &                    \\
		CG             & exact           & WCSS               & simult.          & $\checkmark$   & $\checkmark$   &               &                &                  &                    \\
		CPRBBA         & exact           & WCSS               & simult.          & $\checkmark$   & $\checkmark$   &               &                &                  &                    \\
		\textbf{PC-SOS-SDP}           & exact           & WCSS            & simult.          & $\checkmark$   & $\checkmark$   &               &                &                  &                    \\ \midrule
		\textbf{COPKM} & cb    & WCSS            & sequent.         & $\checkmark$   & $\checkmark$   &               &                &                  &                    \\
		PCKM           & cb    & WCSS + pen.     & sequent.         &                &                & $\checkmark$  & $\checkmark$   & $\checkmark$     &                    \\
		CVQE           & cb    & WCSS + pen.     & sequent.         &                &                & $\checkmark$  & $\checkmark$   &                  &                    \\
		LCVQE          & cb    & WCSS + pen.     & sequent.         &                &                & $\checkmark$  & $\checkmark$   &                  &                    \\
		\textbf{LCC}   & cb    & WCSS + pen.     & sequent.         & $\checkmark$   & ($\checkmark$) &               &  &                  &                    \\
		BCKM           & cb    & WCSS            & simult.          & $\checkmark$   & $\checkmark$   &               &                &                  &                    \\
		BLPKM          & cb    & WCSS            & simult.          & $\checkmark$   & $\checkmark$   &               &                &                  &                    \\
		BHKM           & cb    & WCSS + pen.     & simult.          &                &                & $\checkmark$  & $\checkmark$   & $\checkmark$     &                    \\ \midrule
		\textbf{CSC}   & gb     & cost of cut        & simult.          & ($\checkmark$) & ($\checkmark$) & $\checkmark$  & $\checkmark$   & $\checkmark$     & $\checkmark$       \\
		\textbf{DILS}  & meta       & WCSS + pen.        & simult.          &                &                & $\checkmark$  & $\checkmark$   &                  &                    \\
		SHADE          & meta       & WCSS + pen.        & simult.          &                &                & $\checkmark$  & $\checkmark$   &                  &                    \\ \midrule
		\textbf{PCCC}  & cb    & WCSS + pen.     & simult.          & $\checkmark$   & $\checkmark$   & $\checkmark$  & $\checkmark$   & $\checkmark$     & $\checkmark$       \\ \midrule
		\multicolumn{10}{l}{cb=center-based, gb=graph-based, meta=metaheuristic, WCSS=within cluster sum of squares} \\		\\
	\end{tabularx}
	\caption{Overview of algorithms for clustering with must-link (ML) and cannot-link (CL) constraints\label{tab_literature}.}
\end{table*}

\subsection{Exact algorithms}\label{sec_literature_exact_approaches}
Thanks to algorithmic advances and improvements in hardware (see e.g.~\citealt{bertsimas2017optimal}), exact approaches are increasingly being used to optimally solve small instances of constrained clustering problems. \cite{dao2013declarative} propose a constraint programming approach (CP) that allows to choose between different clustering objectives and integrates various types of constraints, including hard must-link and cannot-link constraints. \cite{babaki2014constrained} introduce a column generation approach (CG) for the minimum within-cluster sum of squares (WCSS) problem with must-link, cannot-link and other additional constraints. \cite{guns2016repetitive} introduce the CPRBBA algorithm, which combines the branch-and-bound algorithm of \cite{brusco2006repetitive} for unconstrained minimum sum-of-squares clustering with a constraint-programming framework to incorporate user constraints. The branch-and-bound algorithm of the CPRBBA is based on computing a sequence of tight lower bounds by evaluating assignments for subsets of objects of increasing size. This idea was first introduced by \cite{koontz1975branch}. \cite{piccialli2022exact} extend an exact solver for the minimum sum-of-squares clustering problem (see \citealt{piccialli2022sos}) to consider hard must-link and cannot-link constraints. Their algorithm is referred to as \textbf{PC-SOS-SDP}. The above-mentioned exact approaches do not consider soft constraints and are only applicable to small instances with up to around 1,000 objects. In our computational experiments, we used the exact approach (PC-SOS-SDP) to determine optimal solutions for small instances. This allows us to evaluate the quality of the solutions produced by our proposed algorithm using the sum-of-squares criterion. 

\subsection{Center-based algorithms}\label{sec_literature_center_based}
Center-based algorithms identify a center for each cluster and aim to minimize the distances between the center of a cluster and the objects that belong to this cluster. The following algorithms are all variants of the well-known k-means algorithm for standard clustering and thus alternate between an object assignment and a cluster center update step until a termination criterion is reached. 

The COP-kmeans algorithm of \cite{wagstaff2001constrained} (\textbf{COPKM}) incorporates the must-link and cannot-link constraints as hard constraints in the assignment step. It assigns objects sequentially to the closest feasible cluster center and stops without a solution if an object cannot be assigned without violating a constraint. The pairwise-constrained k-means algorithm (PCKM) of \cite{basu2004active} is similar to COPKM, but incorporates the pairwise constraints as soft constraints. 
\cite{davidson2005clustering} introduce the constrained vector quantization algorithm (CVQE) algorithm, which iterates through the pairwise constraints in the assignment step rather than the objects and evaluates for each constraint all combinations of assigning the two respective objects. 
The linear CVQE algorithm (LCVQE) of \cite{pelleg2007k} is a modification of the CVQE algorithm with reduced computational complexity. 
The Lagrangian Constrained Clustering (\textbf{LCC}) algorithm of \cite{ganji2016lagrangian} uses a Lagrangian relaxation scheme to iteratively reassign objects that violate pairwise constraints. The algorithm is intended for hard must-link and cannot-link constraints but it does not guarantee to satisfy the hard cannot-link constraints. Therefore, we put a checkmark in parentheses in the column ``Hard CL" in Table~\ref{tab_literature}.
\cite{le2019binary} propose a binary optimization approach for the constrained k-means problem (BCKM), which treats all pairwise constraints as hard constraints and employs an optimization strategy that iteratively updates the cluster center variables and the assignment variables. To update the assignment variables, the binary constraints are relaxed, and a linear program is solved. The solution is projected back to the binary domain using a technique related to the feasibility pump heuristic of \cite{fischetti2005feasibility}. Since the publicly available implementation of the BCKM algorithm can only deal with must-link constraints but not with cannot-link constraints, we did not include this algorithm in our computational analysis. 
The recently introduced binary linear programming-based k-means algorithm (BLPKM) algorithm of \cite{baumann2020binary} assigns objects to clusters by solving a binary linear optimization program in each iteration. The binary linear program includes the must-link and cannot-link constraints as hard constraints. \cite{baumann2022kmeans} present a version of the BLPKM algorithm, referred to as BH-kmeans (BHKM), that includes the must-link and cannot-link constraints as soft constraints. 

To the best of our knowledge, there is no center-based clustering algorithm capable of handling hard and soft must-link and cannot-link constraints simultaneously. In addition, as shown in Table~\ref{tab_literature}, none of the center-based algorithms is able to consider constraint-specific confidence values. Note that the PCCC algorithm that we introduce in this paper can be seen as a combination of the BLPKM and the BH-kmeans algorithms because it inherits the treatment of pairwise constraints as hard constraints from the BLPKM algorithm and the treatment of constraints as soft constraints from the BH-kmeans algorithm. In addition, the PCCC algorithm can account for constraint-specific confidence values and features several methodological enhancements that improve scalability and effectiveness.

\subsection{Non-center based algorithms}\label{sec_literature_other}
In addition to the many center-based approaches, some non-center based approaches were also proposed. \cite{wang2014constrained} propose a spectral clustering-based algorithm (\textbf{CSC}), which allows users to specify must-link and cannot-link constraints with different levels of confidence. The CSC algorithm solves a constrained spectral clustering problem through generalized eigendecomposition such that a user-defined lower bound on the constraint satisfaction is guaranteed. We put the checkmarks in the columns ``Hard ML" and ``Hard CL" in Table~\ref{tab_literature} in parentheses because the algorithm cannot guarantee that individual constraints are satisfied. 
\cite{gonzalez2020dils} introduce the dual iterative local search algorithm (\textbf{DILS}), a metaheuristic that improves an initial solution by applying mutation and recombination operations as well as a local search procedure. The DILS algorithm treats all must-link and cannot-link constraints as soft constraints and thus cannot guarantee that any hard constraints are satisfied. 
\cite{gonzalez2021enhancing} propose another metaheuristic for clustering with soft must-link and cannot-link constraints, the differential evolution-based algorithm (SHADE). Their algorithm achieves similar, although on average slightly worse, ARI values than the DILS algorithm on the benchmark instances used in \cite{gonzalez2020dils}. We therefore do not test it in our computational experiments. As shown in Table~\ref{tab_literature}, also the non-center based algorithms are not able to deal with hard and soft must-link and cannot-link constraints simultaneously. 

\subsection{Search-space reduction techniques}\label{sec_literature_model_size_reduction_techniques} 
Several of the above mentioned algorithms employ specific search-space reduction techniques to increase their efficiency and effectiveness. The branch-and-bound algorithm of \cite{koontz1975branch}, which is refined in the exact constrained clustering approach of \cite{guns2016repetitive}, constructs a sequence of lower bounds that can be used to exclude a large fraction of the search space. To improve the efficiency of the algorithm they propose to divide the data set into subsets and compute bounds for each subset. These bounds can then be combined to obtain tighter bounds when clustering the entire data set. Furthermore a hierarchical scheme for combining subsets is proposed. The sequence in which the subsets are combined is determined based on a distance metric that is computed between the centers of the subset solutions. \cite{guns2016repetitive} propose several improvements to tighten the lower bounds in order to reduce the search space. One improvement is to contract objects subject to must-link constraints into blocks and redefine the model for blocks instead of objects. Further to obtain upper bounds quickly, they fix most variables such that the model greedily adds the best new object to the current partial solution. In addition, to strengthen the lower bounds they propose to consider the full set of constraints when constructing the next partial solution rather than only considering the constraints relevant for the subset of objects considered at this stage. \cite{piccialli2022exact} developed three types of valid inequalities to generate tight bounds and they combine the binary-linear-programming-based k-means algorithm (BLPKM) of \cite{baumann2020binary} with a cluster center initialization technique based on singular value decomposition to obtain feasible upper bounds quickly. 

The computational results presented in the respective papers demonstrate that search-space reduction techniques can significantly enhance algorithm efficiency. The algorithm we introduce in the next section incorporates a search-space reduction technique that introduces only a subset of decision variables such that are large fraction of cannot-link constraints is implicitly satisfied. The resulting mathematical models are considerably smaller than the respective full models, both in terms of number of decision variables and number of cannot-link constraints. The proposed technique differs from the variable fixing technique of \cite{guns2016repetitive} in two important ways. First, the degree of search-space reduction can be controlled by the user, allowing the user to manage the trade-off between computational effort and solution quality. Second, our technique can adjust the size of the search space during the execution of the algorithm by adding more or less decision variables to the mathematical model based on the structure of the current best solution. For example, the penalties associated with violated soft constraints offer crucial information about which assignment variables to include in the mathematical model.

\section{Our PCCC algorithm}\label{sec_algorithm}
\begin{figure*}
	\centering
	\includegraphics[width=\textwidth]{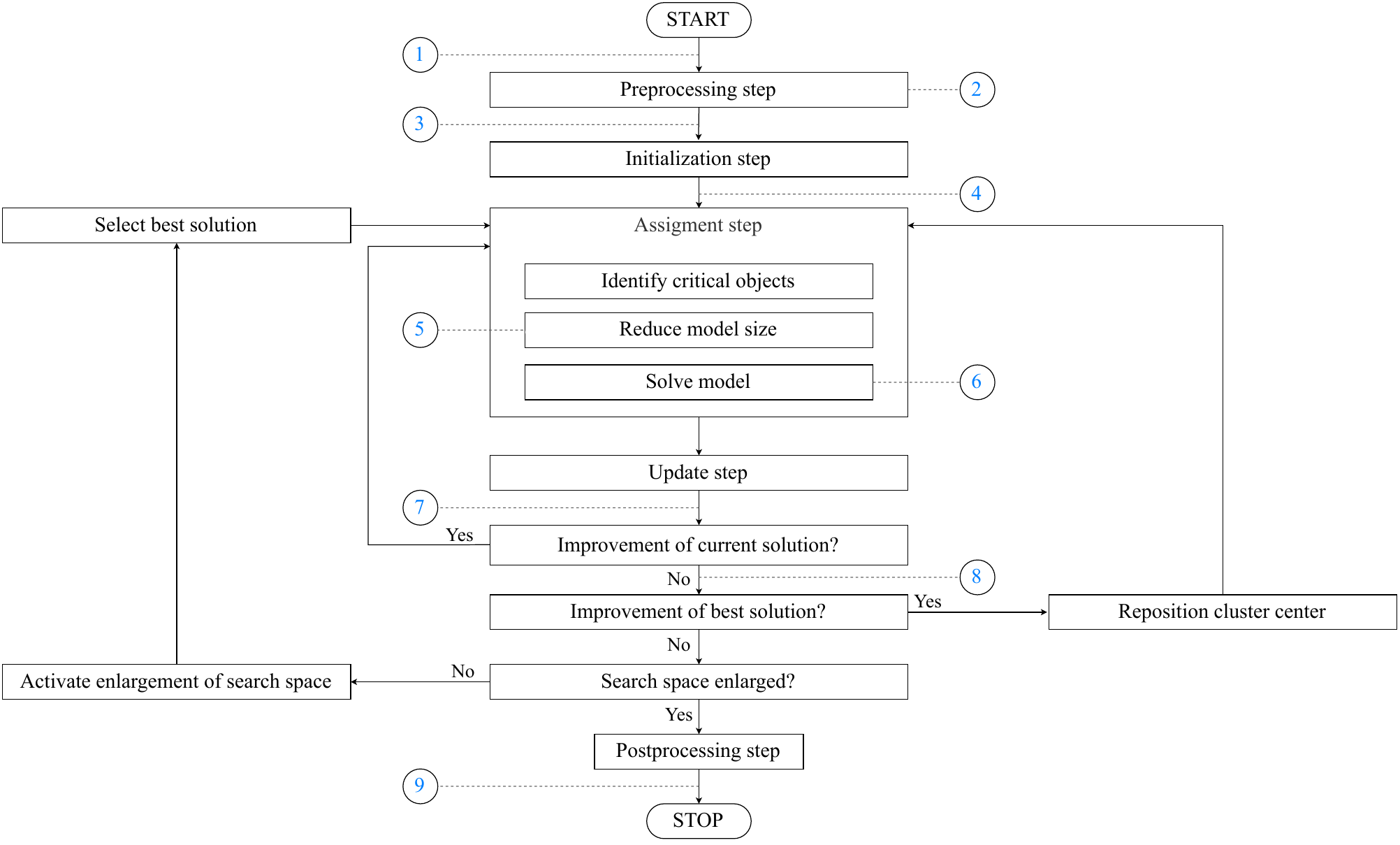}
	\caption{Flowchart of the PCCC algorithm. Illustrations 1--9 are given in Figure~\ref{fig_algorithm_illustrations}}\label{fig_algorithm_flowchart}
\end{figure*}

\begin{figure*}
	\includegraphics[width=\textwidth]{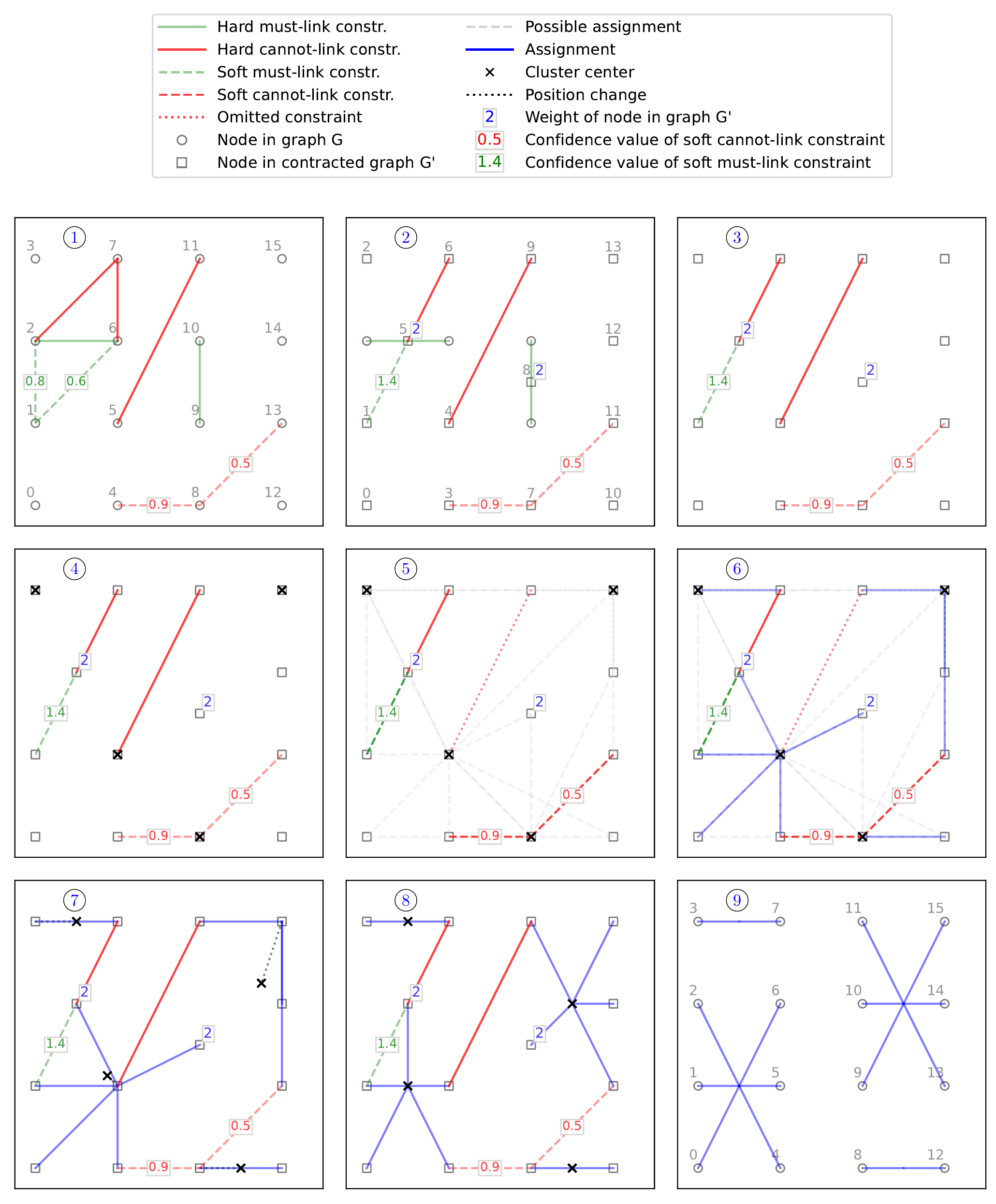}
	\caption{Illustrations for flowchart based on the illustrative example. The algorithm is applied with parameter $q=2$, which explains why the hard cannot-link constraint between the contracted nodes 4 and 9 can be omitted in the assignment step (see illustrations 5 and 6). }\label{fig_algorithm_illustrations}
\end{figure*}

The PCCC algorithm consists of five main steps: preprocessing, initialization, assignment, update, and postprocessing. Figure~\ref{fig_algorithm_flowchart} provides a flowchart of the algorithm. In Sections~\ref{sec_algorithm_preprocessing}--\ref{sec_algorithm_postprocessing}, we describe all five main steps in detail. Figure~\ref{fig_algorithm_illustrations} illustrates the main steps with the illustrative example from Section~\ref{sec_problem_example}. In Sections~\ref{sec_algorithm_repositioninig} and Section~\ref{sec_algorithm_dynamic_enlargement}, we describe two methodological features of the algorithm that increase its effectiveness. 

The input data consists of a set of $n$ objects, each is represented by a $d$-dimensional numeric feature vector $\vect{x}_i \in \mathbb{R}^d$, a parameter $k$ that denotes the number of clusters to be identified, and the additional information consisting of the sets of pairwise constraints $ML$, $CL$, $SML$, $SCL$, and the associated confidence values (see Section~\ref{sec_problem_description}). The input data is represented as a weighted undirected graph $G = (V, E)$ where the node set $V$ corresponds to the objects, and the edge set $E$ corresponds to the pairwise constraints. We distinguish four types of edges: the edges $E^{\textup{ML}}$ correspond to the hard must-link constraints from set $ML$, the edges $E^{\textup{CL}}$ correspond to the hard cannot-link constraints from set $CL$, the edges $E^{\textup{SML}}$ correspond to the soft must-link constraints from set $SML$, and the edges $E^{\textup{SCL}}$ correspond to the soft cannot-link constraints from set $SCL$. The edges representing soft constraints are associated with a confidence value $w_{ij}$. Table~\ref{tab_graphnotation} provides the graph notation that we use in the subsequent sections.

\begin{table*}
	\begin{tabularx}{\textwidth}{lX} \toprule
		\multicolumn{2}{l}{\textbf{Input data}}                                                 \\ \midrule		
		$G=(V, E)$ & Weighted undirected graph that represents the input data\\
		$V$ & Nodes representing objects \\
		$E$ & Edges representing additional information\\
		$E^{\textup{ML}} \subseteq E$ & Edges representing hard must-link constraints \\
		$E^{\textup{CL}} \subseteq E$ & Edges representing hard cannot-link constraints \\		
		$E^{\textup{SML}}\subseteq E$ & Edges representing soft must-link constraints \\						
		$E^{\textup{SCL}}\subseteq E$ & Edges representing soft cannot-link constraints \\
		$w_{ij}$ & Weight of edge $\{i, j\} \in E^{\textup{SML}} \cup E^{\textup{SCL}}$ \\ 						
		$k$ & Number of clusters to be identified \\ 
		$\vect{x}_i \in \mathbb{R}^d$ & Feature vector of node $i$ \\ \midrule
		\multicolumn{2}{l}{\textbf{Input data after preprocessing}}                                                 \\ \midrule				
		$G'=(V', E')$ & Weighted undirected graph after node contraction \\
		$V'$ & Nodes in contracted graph \\
		$E'$ & Edges in contracted graph \\
		$E'^{\textup{CL}} \subseteq E'$ & Edges representing hard cannot-link constraints in contracted graph \\		
		$E'^{\textup{SML}}\subseteq E'$ & Edges representing soft must-link constraints in contracted graph \\						
		$E'^{\textup{SCL}}\subseteq E'$ & Edges representing soft cannot-link constraints in contracted graph\\				
		$m(i)$ & Mapping of a node $i \in V$ to a node in $V'$ \\
		$w'_{ij}$ & Weight of edge $(i, j) \in E'^{\textup{SML}} \cup E'^{\textup{SCL}}$ \\		
		$s_i$ & Weight of node $i$ \\
		$\vect{\bar{x}}_i \in \mathbb{R}^d$ & Feature vector of node $i \in V'$ \\
		\bottomrule
	\end{tabularx}
	\caption{Description of graph notation\label{tab_graphnotation}}
\end{table*}

\subsection{Preprocessing step}\label{sec_algorithm_preprocessing}
In the preprocessing step, we transform graph $G$ into a new graph $G' = (V', E')$ by contracting all edges $\{i, j\} \in E^{\textup{ML}}$ which represent hard must-link constraints. These edge contractions merge each group of nodes that is directly or indirectly connected by hard must-link edges into a single node. Let set $V_i$ denote the nodes of graph $G$ that are merged into node $i$ of the new graph $G'$. The mapping $m: V \rightarrow V'$ indicates for each node in graph $G$ the corresponding node in the new graph. Each node $i \in V'$ is associated with a weight $s_i$ that indicates how many nodes of graph $G$ were merged into this node. The feature vector $\vect{\bar{x}}_i$ of node $i \in V'$ is computed as $\vect{\bar{x}}_i = \frac{\sum_{j \in V_i}\vect{x}_j}{s_i}$. The edge set $E'$ of the new graph $G'$ consists of edges representing hard cannot-link constraints $E'^{\textup{CL}}=\{\{m(i), m(j)\}|\{i, j\} \in E^{\textup{CL}}\}$, edges representing soft must-link constraints $E'^{\textup{SML}}=\{\{m(i), m(j)\}|\{i, j\} \in E^{\textup{SML}}\wedge m(i) \neq m(j)\}$, and edges representing soft cannot-link constraints $E'^{\textup{SCL}}=\{\{m(i), m(j)\}|\{i, j\} \in E^{\textup{SCL}} \wedge m(i) \neq m(j)\}$. In case $E'^{CL}$ contains edges $\{i, j\}$ with $i = j$, then we know that there is no feasible assignment. The weights $w'_{ij}$ of the edges $\{i, j\} \in E'^{\textup{SML}}$ are computed as $w'_{ij} = \sum_{\{i', j'\} \in E^{\textup{SML}} | m(i') = i \wedge m(j') = j}w_{i'j'}$. The weights $w'_{ij}$ of the edges $\{i, j\} \in E'^{\textup{SCL}}$ are computed as $w'_{ij} = \sum_{\{i', j'\} \in E^{\textup{SCL}} | m(i') = i \wedge m(j') = j}w_{i'j'}$. If a pair of nodes $\{i, j\}$ is contained in both edge sets $E'^{SML}$ and $E'^{SCL}$, then we remove the edge with the smaller weight and adjust the other weight by subtracting the smaller weight from it. The output of the preprocessing step is the transformed input data represented as graph $G'$. The idea of reducing the size of the graph by contracting nodes connected by hard must-link constraints has been previously used in other work (see \citealt{guns2016repetitive, ganji2016lagrangian, piccialli2022exact}), but has not yet been explored in the context of soft must-link and cannot-link constraints. 

\subsection{Initialization step}\label{sec_algorithm_initialization}
In the initialization step, we determine the initial positions of the $k$ cluster centers. We provide two different methods for this step. The first method randomly selects $k$ distinct nodes in graph $G'$ and uses their feature vectors $\vect{\bar{x}}_i$ as initial positions. This method is fast but might select some initial positions close to each other so that convergence will be slow. The second method is the k-means++ algorithm of \cite{arthur2006k}. The k-means++ algorithm requires extra running time but tends to spread out the initial positions, which speeds up convergence. 

\subsection{Assignment step}\label{sec_algorithm_assignment}
In the assignment step, each node in graph $G'$ is assigned to one of the $k$ clusters. We solve the mixed binary linear program (MBLP) presented below to assign the nodes to clusters. Table~\ref{tbl_notation} describes the notation used in the formulation.

\begin{table*}
	\begin{tabularx}{\textwidth}{lX}
		\toprule
		\multicolumn{2}{l}{\textbf{Additional parameters}}                                                 \\ \midrule
		$d_{il}$  & Squared Euclidean distance between node~$i \in V'$ and the center of cluster~$l$ \\
		$P$       & Penalty value (by default: $P=\sum_{i \in V';~l=1\ldots,k}d_{il}/(|V'|k)$)					\\ \midrule
		\multicolumn{2}{l}{\textbf{Binary decision variables}}                                  \\ \midrule
		$x_{il}$  & =1, if node~$i \in V'$ is assigned to cluster $l$;~=0, otherwise         \\ \midrule
		\multicolumn{2}{l}{\textbf{Continuous decision variables}}                              \\ \midrule
		$y_{ij}$ & Auxiliary variables to capture violations of soft cannot-link constraint    \\
		$z_{ij}$ & Auxiliary variables to capture violations of soft must-link constraints     \\ \bottomrule
	\end{tabularx}
	\caption{Additional notation used in formulation (MBLP)}\label{tbl_notation}	
\end{table*}

\begin{table*}
{\renewcommand{\arraystretch}{2}
	\setlength\arraycolsep{20pt}
	\[\textup{(MBLP)}\left\{\begin{array}{lllr}
		\textup{Min.}   & \multicolumn{2}{l}{\displaystyle \sum_{i \in V'}\sum_{l=1}^k s_id_{il}x_{il} + P\left(\sum_{\{i, j\} \in E'^{\textup{SCL}}}w'_{ij}y_{ij} + \sum_{\{i, j\} \in E'^{\textup{SML}}}w'_{ij}z_{ij}\right)} & (1) \\
		\textup{s.t.}   & \displaystyle \sum_{l=1}^kx_{il}=1            & (i \in V')                                                                                                                                            & (2) \\
		\vspace{-0.2cm} & \displaystyle x_{il} + x_{jl} \leq 1          & (\{i, j\} \in E'^{\textup{CL}};~l=1,\ldots,k)                                                                                                         & (3) \\
		\vspace{-0.2cm} & \displaystyle x_{il} + x_{jl} \leq 1 + y_{ij} & (\{i, j\} \in E'^{\textup{SCL}};~l=1,\ldots,k)                                                                                                        & (4) \\
		\vspace{-0.2cm} & \displaystyle x_{il} - x_{jl} \leq z_{ij}     & (\{i, j\} \in E'^{\textup{SML}};~l=1,\ldots,k)                                                                                                        & (5) \\
		\vspace{-0.2cm} & x_{il} \in \{0, 1\}                           & (i \in V';~l=1,\ldots,k)                                                                                                                              & (6) \\
		\vspace{-0.2cm} & y_{ij} \geq 0                                 & \{i, j\} \in E'^{\textup{SCL}}                                                                                                                        & (7) \\
		\vspace{-0.2cm} & z_{ij} \geq 0                                 & \{i, j\} \in E'^{\textup{SML}}                                                                                                                        & (8)
	\end{array}\right.\]} 
\end{table*}

The objective function includes two terms. The first term computes the total weighted squared Euclidean distance between the feature vectors of the nodes $i \in V'$ and the centers of their assigned clusters. The second term computes the total penalty value that results from violating soft constraints. Each violation is weighted by $w'_{ij}$ and multiplied with parameter $P$, which trades off the weighted violations against the weighted distances from the first term in the objective function. If parameter $P$ is specified by the user, its value remains constant throughout the algorithm. Otherwise, we set $P$ as the maximum distance between a node and a cluster center. This maximum distance is recalculated in each iteration to accommodate the changing distances between the nodes and the cluster centers. Constraints~(2) ensure that each node is assigned to exactly one cluster. Constraints~(3) represent the hard cannot-link constraints. Constraints~(4) represent the soft cannot-link constraints. If a pair of nodes $\{i, j\} \in E'^{\textup{SCL}}$ is assigned to the same cluster $l$, then the auxiliary variable $y_{ij}$ is forced to take value one which adds the penalty $Pw'_{ij}$ to the objective function. Constraints~(5) represent the soft must-link constraints. If a pair of nodes $\{i, j\} \in E'^{\textup{SML}}$ is assigned to different clusters, then for one $l=1,\ldots,k$, the left-hand side of constraint (5) will be one, and thus forces the auxiliary variable $z_{ij}$ to take value one which adds the penalty $Pw'_{ij}$ to the objective function. Constraints (6)--(8) define the domains of the decision variables. 

To reduce the size of the formulation when the number of clusters, $k$, is large, we propose a model-size reduction technique that allows to assign a node only to one of the $q$ nearest clusters, where the distance of the node from a cluster is determined by the distance to its center. Without any pairwise constraints, every node is assigned to the nearest cluster in an optimal assignment. With pairwise constraints, the nodes might not be assigned to their nearest center to satisfy the constraints. However, rarely will nodes be assigned to distant centers even in the presence of pairwise constraints. By allowing only assignments between a node $i$ and its $q$ nearest clusters, we can reduce the number of binary decision variables from $|V'|k$ to $|V'|q$. The $q$ nearest clusters can be efficiently determined using kd-trees when the number of features is much smaller than the number of nodes (see \citealt{bentley1975multidimensional}). We refer to the reduced mixed binary linear program as (R($q$)MBLP) as it depends on control parameter $q$. If there are no hard cannot-link constraints, the user can choose parameter $q$ arbitrarily to control the trade-off between model size and flexibility to satisfy pairwise constraints. If there are hard cannot-link constraints, we need to set parameter $q$ sufficiently large to guarantee that we can find a feasible assignment if one exists. Let $H=(V',E'^{\textup{CL}})$ be a subgraph of graph $G'$ that has the same node set $V'$ as graph $G'$, but whose edge set only contains the edges that correspond to hard cannot-link constraints. The expression $\textup{deg}_H(i)$ denotes the degree of node $i$ in subgraph $H$ and $\Delta(H)$ is the maximum degree of the nodes in $H$, i.e., $\Delta(H)=\max_{i \in V'}\textup{deg}_H(i)$. 

\begin{lemma}
	If a feasible assignment exists, we find a feasible assignment by solving model (R($q$)MBLP) with $q\geq\min(1 + \Delta(H), k)$.
\end{lemma}

\textbf{Proof.}
Finding an assignment that satisfies all hard cannot-link constraints corresponds to solving a $k$-coloring problem on graph $H=(V', E'^{\textup{CL}})$, which consists of assigning one of $k$ colors to each node $i \in V'$ such that all adjacent nodes have different colors. A feasible solution using at most $k$ colors is called a $k$-coloring. The chromatic number of graph $H$, denoted as $\chi(H)$, is the smallest number of colors required for a $k$-coloring. Hence, we are guaranteed to find a feasible assignment (if one exists) as long as $q \geq \chi(H)$. From \cite{welsh1967upper}, we know that graph $H$ can be colored with at most $\Delta(H) + 1$ colors. Hence $\Delta(H) + 1$ is an upper bound on $\chi(H)$. Consequently, if a feasible assignment exists, we are guaranteed to find it with $q\geq \Delta(H) + 1$. \hfill $\square$ \\

\begin{table*}
	\begin{tabular}{lp{10.5cm}} \toprule
		\multicolumn{2}{l}{\textbf{Additional notation}} \\ \midrule		
		$G''=(V', E'')$ & Subgraph of graph G' \\
		$E''$ & Edges in subgraph $G''$ \\
		$E''^{\textup{CL}}$ & Edges representing hard cannot-link constraints \\
		$E''^{\textup{SCL}}$ & Edges representing soft cannot-link constraints \\ 
		$E''^{\textup{SML}}$ & Edges representing soft must-link constraints \\ 		
		$L_i$ & Clusters to which node $i$ can be assigned \\
		$V'_l$ & Nodes that can be assigned to cluster $l$ \\
		\bottomrule
	\end{tabular}
	\caption{Additional notation used in formulation (R($q$)MBLP)}\label{tbl_notation2}	
\end{table*}

Next, we present the reduced mixed binary linear program (R($q$)MBLP) that we solve if we apply the model-size reduction technique. Let $L_i$ denote the set of $q$ nearest clusters to node $i$. If a cluster does not belong to the $q$ nearest clusters of any node, we add it to set $L_i$ of its closest node $i$. This guarantees that at least one node can be assigned to each cluster. Some hard and soft cannot-link constraints may become redundant if we require nodes to be assigned to one of their $q$ nearest clusters. For example, if two nodes $\{i, j\} \in E'^{\textup{CL}}$ cannot be assigned to the same cluster because $L_i \cap L_j = \emptyset$, the respective cannot-link constraint does not need to be included in the formulation. We introduce the following two new sets of edges that represent hard and soft cannot-link constraints: $E''^{\textup{CL}} = \{\{i, j\} \in E'^{\textup{CL}} | L_i \cap L_j \neq \emptyset\}$, $E''^{\textup{SCL}} = \{\{i, j\} \in E^{\textup{SCL}} | L_i \cap L_j \neq \emptyset\}$. These sets represent only hard and soft cannot-link constraints that can still be violated after applying the model-size reduction technique. Some soft must-link constraints may automatically be violated if we require nodes to be assigned to one of their $q$ nearest clusters. We introduce the following set of soft must-link pairs: $E''^{\textup{SML}} = \{\{i, j\} \in E'^{\textup{SML}} | L_j \cap L_{j'} \neq \emptyset\}$. This set only contains the soft must-link constraints that can potentially be satisfied after applying the model-size reduction technique. Table~\ref{tbl_notation2} describes the additional notation used in the reduced mixed binary linear program (R($q$)MBLP), which reads as follows.

\begin{table*}[h]
{\renewcommand{\arraystretch}{2}
	\setlength\arraycolsep{15pt}
	\[\textup{(R($q$)MBLP)}\left\{\begin{array}{lllr}
		\textup{Min.}   & \multicolumn{2}{l}{\displaystyle \sum_{i \in V'}\sum_{l \in L_i} s_id_{il}x_{il} + P\left(\sum_{\{i, j\} \in E''^{\textup{SCL}}}w'_{ij}y_{ij} + \sum_{\{i, j\} \in E''^{\textup{SML}}}w'_{ij}z_{ij}\right)} & (9)  \\
		\textup{s.t.}   & \displaystyle \sum_{l \in L_i}x_{il}=1       & (i \in V')                                                                                                                                                   & (10)  \\
		                & x_{il} + x_{jl} \leq 1                       & (\{i, j\} \in E''^{\textup{CL}};~l \in L_i \cap L_j)                                                                                                         & (11)  \\
		                & x_{il} + x_{jl} \leq 1 + y_{ij}              & (\{i, j\} \in E''^{\textup{SCL}};~l \in L_i \cap L_j)                                                                                                        & (12)  \\
		                & x_{il} - x_{jl} \leq z_{ij}                  & (\{i, j\} \in E''^{\textup{SML}};~l \in L_i \cap L_j)                                                                                                        & (13)  \\
		                & x_{il} \leq z_{ij}                           & (\{i, j\} \in E''^{\textup{SML}};~l \in L_i \setminus L_j)                                                                                                   & (14)  \\
		                & x_{jl} \leq z_{ij}                           & (\{i, j\} \in E''^{\textup{SML}};~l \in L_j \setminus L_i)                                                                                                   & (15)  \\
		\vspace{-0.2cm} & x_{il} \in \{0, 1\}                          & (i \in V';~l \in L_i)                                                                                                                                        & (16)  \\
		\vspace{-0.2cm} & y_{ij} \geq 0                                & \{i, j\} \in E''^{\textup{SCL}}                                                                                                                              & (17) \\
		\vspace{-0.2cm} & z_{ij} \geq 0                                & \{i, j\} \in E''^{\textup{SML}}                                                                                                                              & (18)
	\end{array}\right.\]}
\end{table*}

Most of the constraints of the model (R($q$)MBLP) are similar to the constraints of the model (MBLP) and are thus not further explained here. Constraints~(14) are required to ensure that $z_{ij}$ takes value one if node~$i$ is assigned to a cluster $l$ that is not among the $q$ nearest centers of node~$j$. Analogously, constraints~(15) ensure that $z_{ij}$ takes value one if node~$j$ is assigned to a cluster $l$ that is not among the $q$ nearest centers of node~$i$. Note that when using model (R($q$)MBLP), parameter $P$ is computed as $P=\sum_{i \in V';~l \in L_i}d_{il}/(|V'|q)$ by default.

\subsection{Update step}\label{sec_algorithm_update}
In the update step, the positions of the cluster centers are adjusted based on the assignment determined in the previous step. We denote the positions of the cluster centers as $\vect{\bar{o}}_l$. Let $A_l$ be the set of nodes assigned to cluster $l$. The position of the center of cluster $l$ is then computed as $\vect{\bar{o}}_l = \frac{\sum_{i \in A_l}s_i\vect{\bar{x}}_i}{\sum_{i \in A_l}s_i}$. If a cluster center has no assignments, we reposition it as follows. We lexicographically rank all clusters based on two criteria: first, the total penalties of the cluster members involved in violations of soft cannot-link constraints, and second, the cluster Inertia, both in descending order. From the cluster that ranks highest, we randomly select one object. The feature values of this selected object are then used to determine the coordinates of the new cluster position. To ensure that all clusters contain at least one object, the selected object is assigned to the respective cluster center. The assignment and the update steps are repeated as long as the objective function value of the (reduced) mixed binary linear program (R($q$)MBLP), computed after the update step, can be decreased. The best assignment at termination is passed to the postprocessing step. 

\subsection{Postprocessing step}\label{sec_algorithm_postprocessing}
In the postprocessing step, we map the labels of the nodes $i \in V'$ back to the original objects and return the final assignment to the user.

\subsection{Cluster repositioning}\label{sec_algorithm_repositioninig}
\begin{figure*}
	\centering
	\includegraphics[width=\textwidth]{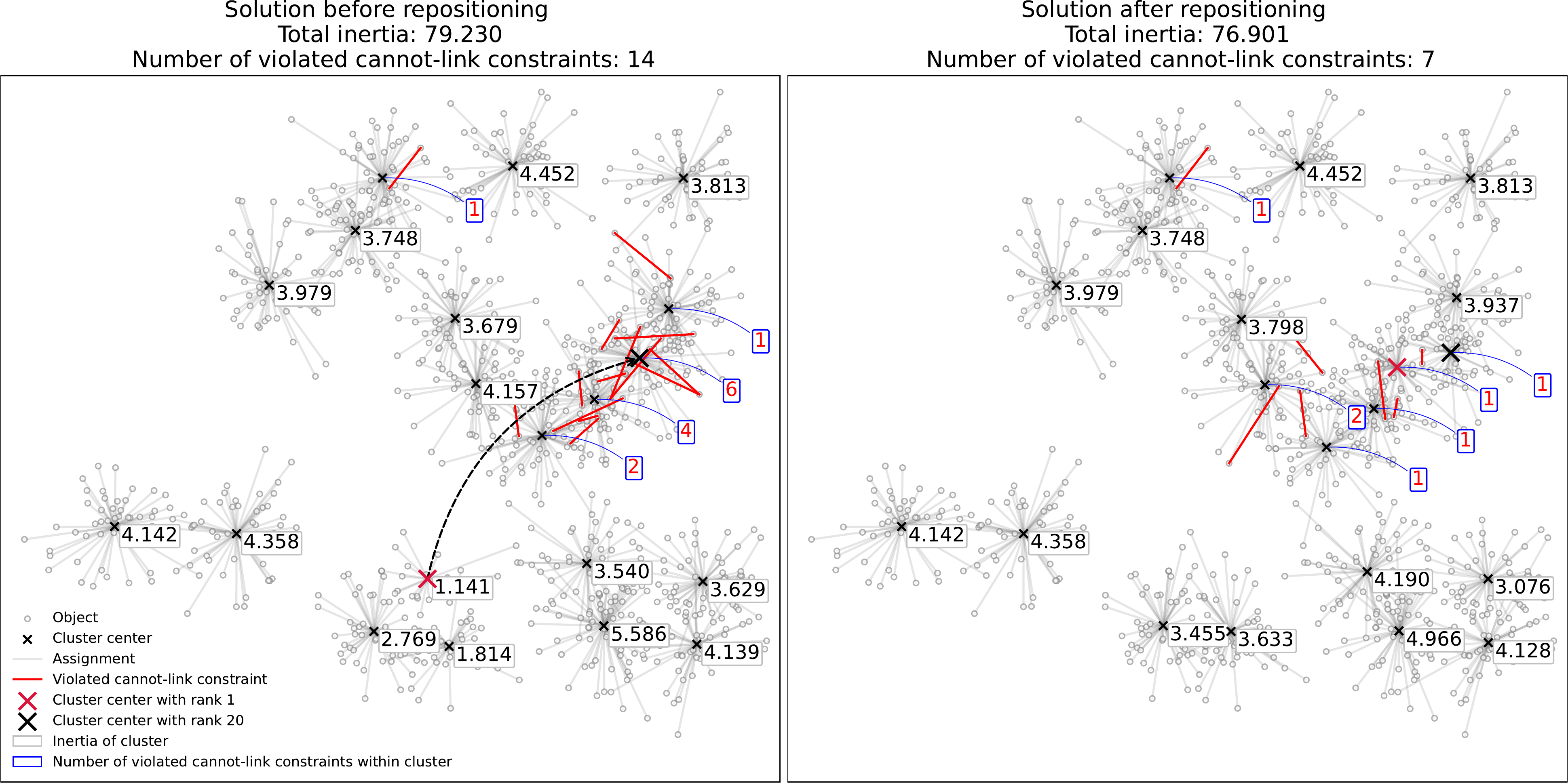}
	\caption{Illustration of cluster repositioning with the synthetic data set n1000-k20 and the cannot-link constraints of constraint set 15\%~CS (provided as soft constraints). The left plot shows the converged solution with $q=2$ before the repositioning and the plot on the right shows the converged solution with $q=2$ after repositioning. Both the number of violations of cannot-link constraints as well as the total Inertia could be improved by the repositioning. }\label{fig_algorithm_repositioning}
\end{figure*}

The PCCC algorithm is a randomized algorithm because it uses the k-means++ algorithm to select the initial cluster center positions. The common practice for randomized algorithms is to restart them several times, each time with a different random seed. However, this common practice misses the opportunity to extract useful information from the previous solution to initialize the next solution. We introduce here a new concept that extracts information from the converged solution, adjusts it and continues with the adjusted solution. We refer to this concept that is applicable to any multi-start method as \textit{enhanced multi-start}. The implementation works as follows. When the objective function value is no longer improved, we reposition one cluster center, reset the objective function value of the current solution to a large value and solve the assignment step with the modified cluster centers. The idea is to identify areas in the feature space where an additional cluster center could help reduce the penalties from violation of soft constraints or help reduce the Inertia of the nearby clusters. We lexicographically rank all clusters based on two criteria: first, the total penalties of the cluster members involved in violations of soft cannot-link constraints, and second, the total cluster Inertia, both in descending order. We then set the position of the cluster that ranks lowest (no or few violations and low Inertia) to the position of the cluster center that ranks highest (many violations and high Inertia). Figure~\ref{fig_algorithm_repositioning} illustrates the cluster repositioning for the synthetic data set n1000-k20 from COL3 with 1,000 objects and 20 clusters and the cannot-link constraints (provided as soft constraints) from the corresponding constraint set 15\%~CS. In order to simplify the presentation, we did not use the must-link constraints of constraint set 15\%~CS here. The left subplot shows a solution obtained with $q=2$ that cannot be further improved. Five clusters in this solution contain objects involved in the violation of soft cannot-link constraints. These violated constraints are highlighted with red lines in the plot and the red numbers indicate the number of violated constraints in the respective clusters. For the clusters with no violations, we state the Inertia value next to the cluster center. After the lexicographical sorting, the cluster with Inertia value 1.141 ranks first and is set to the position of the cluster that ranks last, i.e., the one with 6 violations. After this repositioning, the assignment and update steps are repeated with the new cluster center positions until the solution can no longer be improved. The plot on the right shows the converged solution which has considerably fewer violations and a lower total Inertia value. 

\subsection{Dynamic enlargement of the search space}\label{sec_algorithm_dynamic_enlargement} 
\begin{figure*}
	\centering
	\includegraphics[width=\textwidth]{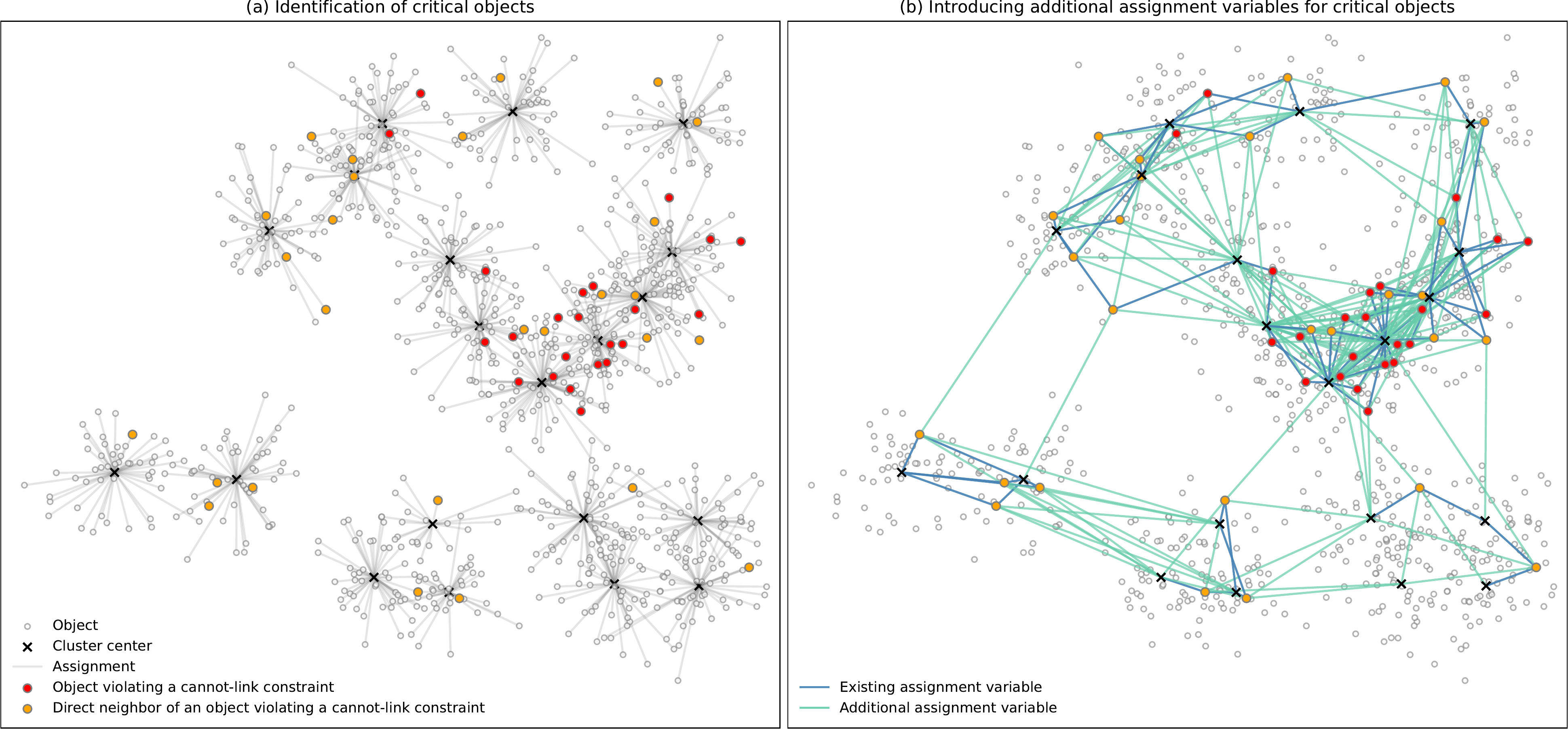}
	\caption{Illustration of the dynamic enlargement of the search space with the synthetic data set n1000-k20 and the cannot-link constraints of constraint set 15\%~CS (provided as soft constraints). The left plot highlights the $\gamma=50$ critical objects (red and orange) in the converged solution obtained with $q=2$. The right plot highlights the additional $\delta=3$ assignment variables that are introduced for the critical objects. With the enlarged search space (150 additional binary variables) all violations can be resolved in the next assignment step. Increasing $q$ uniformly for all objects from $q=2$ to $q=5$ would have let to an increase of 3,000 binary variables. }\label{fig_algorithm_enlargement}
\end{figure*}

As previously mentioned, the parameter $q$ controls the balance between computational efficiency and solution quality by allowing assignments of objects solely to the $q \ll k$ closest cluster centers. However, if the number of objects $n$ is very large, even a small uniform increase in $q$ for all objects can significantly slow down the solver. To address this issue, we propose to increase $q$ by $\delta$ not uniformly for all objects, but only for a user-defined number $\gamma$ of so-called critical objects. The identification of critical objects works as follows. We first identify objects that are involved in violations of soft cannot-link constraints and sort them according to the respective total penalty associated with the violation of soft cannot-link constraints in descending order. The first $\gamma$ objects are considered critical. If fewer than $\gamma$ objects are involved in violations, we randomly add objects that are directly connected to the critical objects via a soft cannot-link constraint. The underlying idea is that to resolve violations, the solver sometimes has to consider chains of objects connected by soft cannot-link constraints. If after adding additional objects, the number of critical objects is still smaller than $\gamma$, we randomly add any of the remaining objects. Model (R($q$)MBLP) is then solved with the adjusted sets $L_i$. Note that in order to get high-quality solutions quickly, the dynamic enlargement of search space is only activated after the best solution can no longer be improved (see~Figure~\ref{fig_algorithm_flowchart}). The dynamic enlargement of the search space is illustrated in Figure~\ref{fig_algorithm_enlargement}.

\section{Computational experiments}\label{sec_experiment}
In this section, we compare the PCCC algorithm with state-of-the-art approaches and examine how the complexity parameters of instances affect its performance. Since the state-of-the-art approaches are less general in their applicability than the PCCC algorithm, we focus on instances to which all approaches are applicable. In Section~\ref{sec_experiments_benchmark_approaches}, we list the state-of-the-art approaches. In Section~\ref{sec_experiments_data_sets}, we introduce the four collections of data sets that we use in the computational experiments. In Section~\ref{sec_experiments_constraint_sets}, we describe the constraint sets that accompany the data sets. In Section~\ref{sec_experiments_optimal_solutions} we compare the solutions produced by the PCCC algorithm to optimal solutions. In Sections~\ref{sec_experiments_comparison_col1}--\ref{sec_experiments_comparison_col4}, we report and discuss the results of the algorithms for the four different data set collections, respectively. All experiments were executed on an HP workstation with two Intel Xeon CPUs with clock speed 3.30 GHz and 256 GB of RAM. The code of the PCCC algorithm is publicly available on \href{\githubalgorithm}{\color{blue}GitHub} (see \citealt{bauhoc2024pccc}).

\subsection{State-of-the-art approaches}\label{sec_experiments_benchmark_approaches}
We use the state-of-the-art solver PC-SOS-SDP of \cite{piccialli2022exact} to generate optimal solutions for small instances. The DILS algorithm of \cite{gonzalez2020dils} represents the current state-of-the-art in terms of the Adjusted Rand Index (ARI). In a recent computational comparison, it outperformed six established algorithms for clustering with must-link and cannot-link constraints (see \citealt{gonzalez2020dils}). We, therefore, include the DILS algorithm in our experiments. In addition, we include the Lagrangian Constrained Clustering (LCC) algorithm of \cite{ganji2016lagrangian} and the spectral clustering-based algorithm (CSC) of \cite{wang2014constrained} as they have not yet been compared to the DILS algorithm in the literature. We include the COP-Kmeans algorithm (COPKM) of \cite{wagstaff2001constrained} because it is probably the most popular algorithm for clustering with must-link and cannot-link constraints. Finally, to assess the impact of considering pairwise constraints, we also include the standard k-means algorithm, which disregards all pairwise constraints. Table~8 in Appendix~A of the Online Supplement lists for each tested approach the corresponding paper and the programming language of the implementation.

\subsection{Data sets}\label{sec_experiments_data_sets}
We use four collections of data sets, which we refer to as COL1, COL2, COL3, and COL4. All data sets are classification data sets, i.e., a ground truth assignment is available to evaluate the clustering solution with an external performance measure such as the Adjusted Rand Index (ARI). Table~\ref{tbl_collections} provides an overview of the four collections. A detailed description of the collections can be found in Appendix~B of the Online Supplement.

\begin{table*}[ht]
	\centering
	\begin{tabularx}{\textwidth}{Xrrrrl}
\toprule
 Collection name & Data sets & Objects & Features & Classes & Type \\
\midrule
COL1 & 25 & 47--846 & 2--90 & 2--15 & Real and syntehtic \\
COL2 & 15 & 300--300 & 2--2 & 10--50 & Syntehtic \\
COL3 & 24 & 500--5,000 & 2--2 & 2--100 & Syntehtic \\
COL4 & 6 & 5,300--70,000 & 2--3,072 & 2--100 & Real and syntehtic \\
\bottomrule
\end{tabularx}

	\caption{The table reports for each collection the number of data sets, the minimum and maximum size of a data set in the collection in terms of number of objects, number of features, and number of classes (here clusters), and the type}\label{tbl_collections}
\end{table*}

\subsection{Constraint sets}\label{sec_experiments_constraint_sets}
For each data set, we generated noise-free constraint sets of different size with the procedure proposed by \cite{gonzalez2020dils}. Noise-free constraints agree with the ground truth labels of a data set, i.e., if two objects are subject to a must-link constraint, then they have the same ground truth label. If two objects are subject to a cannot-link constraint, they have different ground truth labels. A detailed description of how we generated the noise-free constraint sets can be found in Appendix~C of the Online Supplement. On \href{\githubresults}{\color{blue}GitHub}, Tables~W1--W4 list the number of must-link and cannot-link constraints in the noise-free constraint sets of the collections COL1--COL4, respectively. The tables show that the ratio between the number of must-link and the number of cannot-link constraints varies greatly between constraint sets. The constraint sets associated with data sets with many classes tend to have more cannot-link constraints. Instances with constraint sets that mostly contain cannot-link constraints are considered more challenging than instances with constraint sets that mostly contain must-link constraints. We provide a code on \href{\githubdata}{\color{blue}GitHub} for making all collections and the constraint sets available. Running this code downloads the publicly available data sets and constraint sets and generates the synthetic data sets and constraint sets we introduce in this paper. 

\subsection{Comparison to optimal solutions}\label{sec_experiments_optimal_solutions} 
\begin{figure*}
	\includegraphics[width=0.49\textwidth]{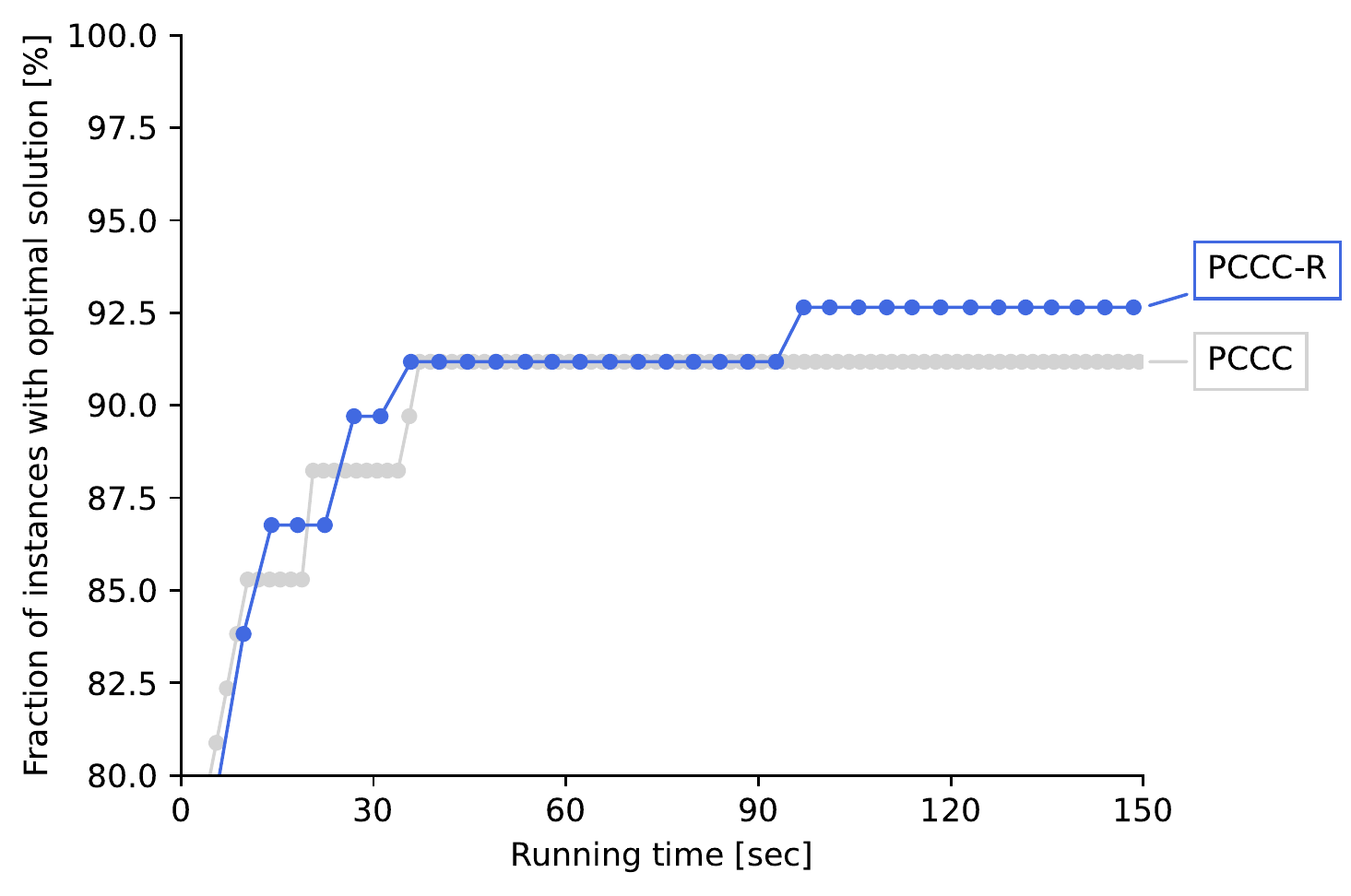}\hfill\includegraphics[width=0.49\textwidth]{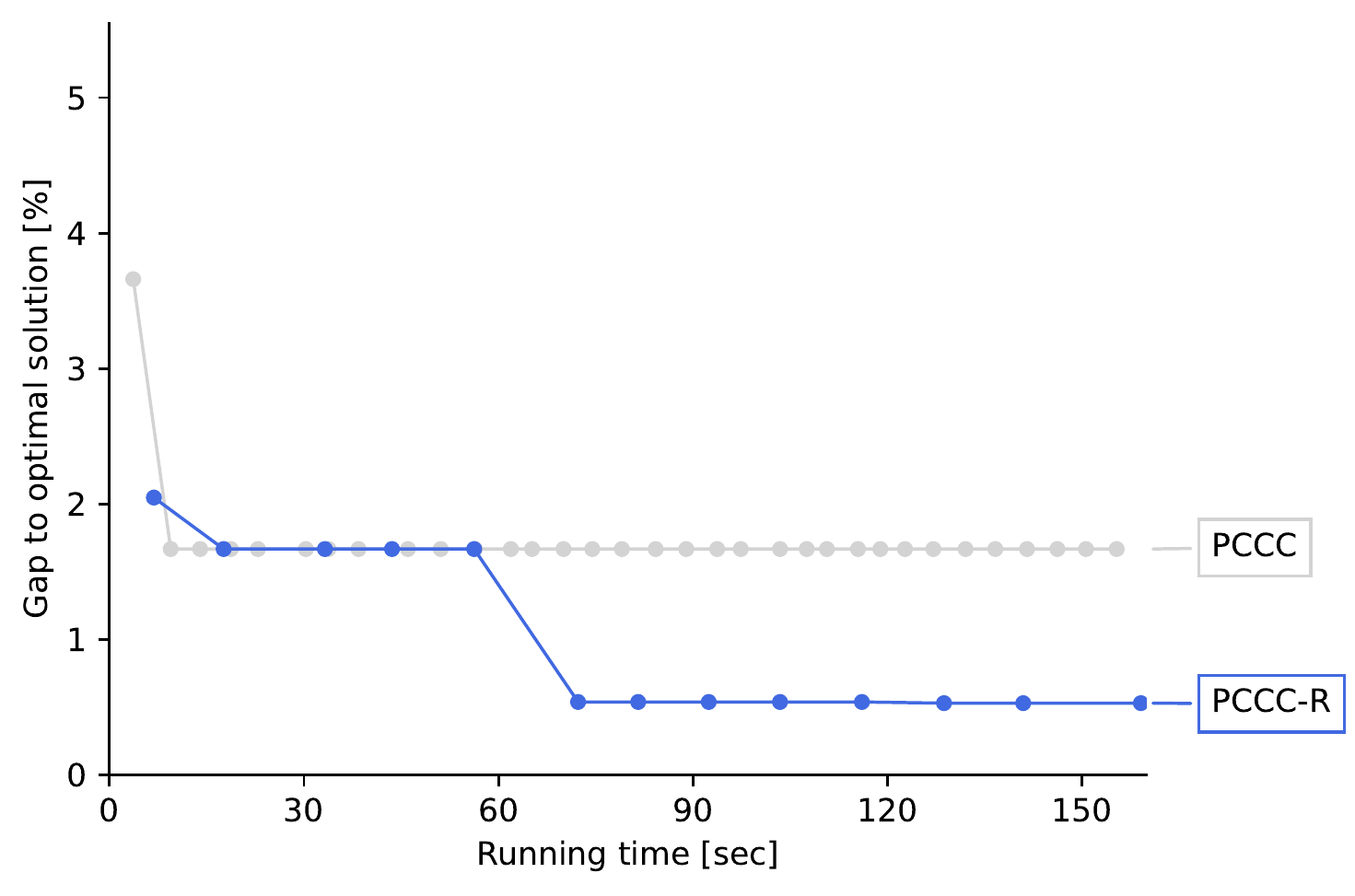}
	\caption{The left plot shows how the fraction of optimal solutions found increases with increasing the number of repetitions (running time) of the algorithms. The right plot shows for the data set Movement Libras and constraint set size 5\% CS how the gap to the optimal solution decreases with increasing repetitions (running time). The PCCC-R algorithm obtains lower gaps earlier than the PCCC algorithm for data sets with a large number of clusters, which demonstrates the effectiveness of the cluster repositioning. \label{fig_inertia_comparison_opt}}
\end{figure*}

We first compare the solutions produced by the PCCC algorithm to optimal solutions that we determined using the state-of-the-art solver for minimum sum-of-squares clustering (PC-SOS-SDP) of \cite{piccialli2022exact}. For this comparison, we used the 25 well-known benchmark data sets and the corresponding constraint sets from collection COL1. Since there are four constraint sets for each data set, the test set for this comparison comprises 100 problem instances. We applied the PC-SOS-SDP solver with a time limit of 3,600 seconds to each instance. The solver was able to prove optimality for 68 out of 100 instances within this time limit. Especially with a large number of CL constraints, the solver was often not able to prove optimality within the time limit. We then applied the PCCC algorithm (PCCC) and the PCCC algorithm with cluster repositioning (PCCC-R) 3 times to each of the 68 instances, every time with a different random seed and a time limit of 1,800 seconds. In Appendix~D of the Online Supplement, we list the settings used to run the algorithms. Table~\ref{tbl_inertia_experiment1_optimal_solutions} lists all 68 instances for which an optimal solution (Opt.\ Gap [\%]=0.00) was found within the time limit and reports for each of them the optimal Inertia value in column 3 and the required running time of the PC-SOS-SDP solver in column 5. Columns 6, 7, and 8 contain the results of the PCCC algorithm. Specifically, the three columns report for each instance the minimum Inertia value (from the 3 repetitions), the relative gap between the minimum Inertia value and the optimal Inertia value, and the running time of the PCCC algorithm, respectively. Columns 9, 10, and 11 report the same information for the PCCC-R algorithm. The last row of the table provides the average values for the different columns. Overall, the table indicates that the solutions devised by the two PCCC variants are of high quality. With only 3 repetitions, the PCCC-R algorithm produced 49 optimal solutions (72.0\%), while the PCCC algorithm produced 45 optimal solutions (66.2\%). The maximum relative gap is only 1.7\% for the PCCC-R algorithms and 5.13\% for the PCCC algorithm. As shown in the left plot of Figure~\ref{fig_inertia_comparison_opt}, the fraction of optimal solutions found by the PCCC-R and PCCCC algorithms can be increased by increasing the number of repetitions, i.e., the running time. The right subplot of Figure~\ref{fig_inertia_comparison_opt} visualizes for the Movement Libras data set (which has the largest number of clusters) and constraint set CS 5\% how the average relative gap to the optimal solutions decreases as the number of repetitions (or running time) is increased. The plot demonstrates that for data sets with many clusters, the cluster repositioning is particularly effective. 

\begin{table*}
	\centering
	\tiny
	\renewcommand{\tabcolsep}{4pt}
	\begin{tabularx}{\textwidth}{Xrrrrrrrrrr}
\toprule
\multicolumn{2}{c}{} & \multicolumn{3}{c}{PC-SOS-SDP-Solver} & \multicolumn{3}{c}{PCCC} & \multicolumn{3}{c}{PCCC-R} \\ \cmidrule(lr){3-5} \cmidrule(lr){6-8} \cmidrule(lr){9-11}\
Dataset & Constraint set [\%] & Inertia & Opt. Gap [\%] & Running time [s] & Inertia & Opt. Gap [\%] & Running time [s] & Inertia & Opt. Gap [\%] & Running time [s] \\
\midrule
Appendicitis & 5 & \bfseries 491.80 & 0.00 & 387.0 & \bfseries 491.80 & 0.00 & \bfseries 0.3 & \bfseries 491.80 & 0.00 & 0.9 \\
Appendicitis & 10 & \bfseries 544.18 & 0.00 & 116.0 & \bfseries 544.18 & 0.00 & \bfseries 0.3 & \bfseries 544.18 & 0.00 & 0.5 \\
Appendicitis & 15 & \bfseries 612.93 & 0.00 & 32.0 & \bfseries 612.93 & 0.00 & \bfseries 0.1 & \bfseries 612.93 & 0.00 & 0.4 \\
Appendicitis & 20 & \bfseries 612.93 & 0.00 & 28.0 & \bfseries 612.93 & 0.00 & \bfseries 0.1 & \bfseries 612.93 & 0.00 & 0.1 \\
Breast Cancer & 5 & \bfseries 12,084.17 & 0.00 & 131.0 & \bfseries 12,084.17 & 0.00 & \bfseries 0.8 & \bfseries 12,084.17 & 0.00 & 1.8 \\
Breast Cancer & 10 & \bfseries 12,185.03 & 0.00 & 28.0 & \bfseries 12,185.03 & 0.00 & \bfseries 0.2 & \bfseries 12,185.03 & 0.00 & 0.3 \\
Bupa & 5 & \bfseries 1,849.95 & 0.00 & 1,459.0 & \bfseries 1,849.95 & 0.00 & \bfseries 1.1 & \bfseries 1,849.95 & 0.00 & 2.0 \\
Bupa & 10 & \bfseries 2,041.64 & 0.00 & 63.0 & 2,041.65 & 0.00 & \bfseries 0.4 & 2,041.65 & 0.00 & 0.5 \\
Circles & 5 & \bfseries 473.13 & 0.00 & 1,199.0 & 473.83 & 0.15 & \bfseries 0.9 & 473.83 & 0.15 & 2.2 \\
Circles & 10 & \bfseries 598.80 & 0.00 & 35.0 & \bfseries 598.80 & 0.00 & \bfseries 0.4 & \bfseries 598.80 & 0.00 & 0.7 \\
Ecoli & 15 & \bfseries 1,027.69 & 0.00 & 339.0 & 1,028.14 & 0.04 & \bfseries 2.7 & 1,028.14 & 0.04 & 6.3 \\
Ecoli & 20 & \bfseries 1,111.78 & 0.00 & 57.0 & 1,115.66 & 0.35 & \bfseries 0.8 & 1,115.66 & 0.35 & 2.9 \\
Glass & 5 & \bfseries 812.41 & 0.00 & 438.0 & 814.14 & 0.21 & \bfseries 2.2 & 814.14 & 0.21 & 5.0 \\
Glass & 15 & \bfseries 1,266.22 & 0.00 & 998.0 & 1,312.32 & 3.64 & \bfseries 1.4 & \bfseries 1,266.22 & 0.00 & 3.7 \\
Glass & 20 & \bfseries 1,409.89 & 0.00 & 484.0 & 1,410.63 & 0.05 & \bfseries 1.1 & 1,410.63 & 0.05 & 2.2 \\
Haberman & 5 & \bfseries 799.25 & 0.00 & 1,900.0 & 799.64 & 0.05 & \bfseries 0.7 & 799.64 & 0.05 & 2.0 \\
Haberman & 10 & \bfseries 889.43 & 0.00 & 35.0 & 889.45 & 0.00 & \bfseries 0.3 & 889.45 & 0.00 & 0.6 \\
Haberman & 15 & \bfseries 891.42 & 0.00 & 27.0 & \bfseries 891.42 & 0.00 & \bfseries 0.1 & \bfseries 891.42 & 0.00 & 0.2 \\
Hayesroth & 5 & \bfseries 446.67 & 0.00 & 1,682.0 & 446.73 & 0.01 & \bfseries 0.6 & \bfseries 446.67 & 0.00 & 2.0 \\
Hayesroth & 15 & \bfseries 553.10 & 0.00 & 48.0 & \bfseries 553.10 & 0.00 & \bfseries 0.6 & \bfseries 553.10 & 0.00 & 1.1 \\
Hayesroth & 20 & \bfseries 551.07 & 0.00 & 34.0 & \bfseries 551.07 & 0.00 & \bfseries 0.2 & \bfseries 551.07 & 0.00 & 0.5 \\
Heart & 5 & \bfseries 3,044.98 & 0.00 & 405.0 & \bfseries 3,044.98 & 0.00 & \bfseries 0.8 & \bfseries 3,044.98 & 0.00 & 1.5 \\
Heart & 10 & \bfseries 3,085.12 & 0.00 & 33.0 & \bfseries 3,085.12 & 0.00 & \bfseries 0.4 & \bfseries 3,085.12 & 0.00 & 1.0 \\
Ionosphere & 5 & \bfseries 10,259.04 & 0.00 & 2,321.0 & 10,259.72 & 0.01 & \bfseries 0.6 & 10,259.72 & 0.01 & 1.9 \\
Ionosphere & 10 & \bfseries 10,944.91 & 0.00 & 39.0 & \bfseries 10,944.91 & 0.00 & \bfseries 0.3 & \bfseries 10,944.91 & 0.00 & 0.5 \\
Iris & 5 & \bfseries 146.30 & 0.00 & 222.0 & 146.39 & 0.06 & \bfseries 0.7 & \bfseries 146.30 & 0.00 & 1.7 \\
Iris & 10 & \bfseries 146.15 & 0.00 & 110.0 & \bfseries 146.15 & 0.00 & \bfseries 1.2 & \bfseries 146.15 & 0.00 & 1.7 \\
Iris & 15 & \bfseries 140.97 & 0.00 & 212.0 & 141.15 & 0.13 & \bfseries 0.8 & \bfseries 140.97 & 0.00 & 1.7 \\
Iris & 20 & \bfseries 145.01 & 0.00 & 305.0 & 145.02 & 0.01 & \bfseries 0.7 & 145.02 & 0.01 & 1.1 \\
Led7Digit & 15 & \bfseries 1,449.42 & 0.00 & 702.0 & 1,449.62 & 0.01 & \bfseries 16.7 & 1,449.60 & 0.01 & 52.9 \\
Led7Digit & 20 & \bfseries 1,511.03 & 0.00 & 68.0 & \bfseries 1,511.03 & 0.00 & \bfseries 6.7 & \bfseries 1,511.03 & 0.00 & 16.9 \\
Monk2 & 5 & \bfseries 2,358.86 & 0.00 & 481.0 & \bfseries 2,358.86 & 0.00 & \bfseries 1.7 & \bfseries 2,358.86 & 0.00 & 3.5 \\
Monk2 & 10 & \bfseries 2,382.69 & 0.00 & 28.0 & \bfseries 2,382.69 & 0.00 & \bfseries 0.3 & \bfseries 2,382.69 & 0.00 & 0.6 \\
Moons & 5 & \bfseries 283.74 & 0.00 & 1,243.0 & \bfseries 283.74 & 0.00 & \bfseries 0.6 & \bfseries 283.74 & 0.00 & 1.7 \\
Moons & 10 & \bfseries 322.11 & 0.00 & 28.0 & \bfseries 322.11 & 0.00 & \bfseries 0.4 & \bfseries 322.11 & 0.00 & 0.9 \\
Moons & 15 & \bfseries 322.93 & 0.00 & 27.0 & \bfseries 322.93 & 0.00 & \bfseries 0.1 & \bfseries 322.93 & 0.00 & 0.1 \\
Movement Libras & 5 & \bfseries 10,597.86 & 0.00 & 3,309.0 & 10,774.72 & 1.67 & \bfseries 9.4 & 10,774.72 & 1.67 & 17.6 \\
Newthyroid & 5 & \bfseries 506.06 & 0.00 & 696.0 & \bfseries 506.06 & 0.00 & \bfseries 0.8 & \bfseries 506.06 & 0.00 & 2.9 \\
Newthyroid & 10 & \bfseries 536.00 & 0.00 & 255.0 & \bfseries 536.00 & 0.00 & \bfseries 0.4 & \bfseries 536.00 & 0.00 & 1.2 \\
Newthyroid & 15 & \bfseries 550.93 & 0.00 & 32.0 & \bfseries 550.93 & 0.00 & \bfseries 0.2 & \bfseries 550.93 & 0.00 & 0.4 \\
Newthyroid & 20 & \bfseries 550.36 & 0.00 & 34.0 & \bfseries 550.36 & 0.00 & \bfseries 0.4 & \bfseries 550.36 & 0.00 & 0.7 \\
Saheart & 5 & \bfseries 3,756.52 & 0.00 & 941.0 & \bfseries 3,756.52 & 0.00 & \bfseries 0.9 & \bfseries 3,756.52 & 0.00 & 2.7 \\
Saheart & 10 & \bfseries 3,924.31 & 0.00 & 29.0 & \bfseries 3,924.31 & 0.00 & \bfseries 0.3 & \bfseries 3,924.31 & 0.00 & 0.6 \\
Sonar & 5 & \bfseries 11,288.49 & 0.00 & 3,564.0 & 11,294.58 & 0.05 & \bfseries 0.7 & 11,294.58 & 0.05 & 2.8 \\
Sonar & 10 & \bfseries 11,873.84 & 0.00 & 37.0 & \bfseries 11,873.84 & 0.00 & \bfseries 0.5 & \bfseries 11,873.84 & 0.00 & 1.1 \\
Sonar & 15 & \bfseries 11,962.93 & 0.00 & 27.0 & \bfseries 11,962.93 & 0.00 & \bfseries 0.1 & \bfseries 11,962.93 & 0.00 & 0.2 \\
Soybean & 5 & \bfseries 367.14 & 0.00 & 31.0 & \bfseries 367.14 & 0.00 & \bfseries 0.2 & \bfseries 367.14 & 0.00 & 0.5 \\
Soybean & 10 & \bfseries 367.14 & 0.00 & 31.0 & \bfseries 367.14 & 0.00 & \bfseries 0.2 & \bfseries 367.14 & 0.00 & 0.7 \\
Soybean & 15 & \bfseries 367.14 & 0.00 & 27.0 & \bfseries 367.14 & 0.00 & \bfseries 0.3 & \bfseries 367.14 & 0.00 & 0.6 \\
Soybean & 20 & \bfseries 367.14 & 0.00 & 28.0 & \bfseries 367.14 & 0.00 & \bfseries 0.2 & \bfseries 367.14 & 0.00 & 0.6 \\
Spectfheart & 5 & \bfseries 10,467.36 & 0.00 & 1,743.0 & 10,468.04 & 0.01 & \bfseries 0.7 & 10,468.04 & 0.01 & 1.5 \\
Spectfheart & 10 & \bfseries 11,210.45 & 0.00 & 35.0 & \bfseries 11,210.45 & 0.00 & \bfseries 0.4 & \bfseries 11,210.45 & 0.00 & 0.6 \\
Spectfheart & 15 & \bfseries 11,268.27 & 0.00 & 27.0 & \bfseries 11,268.27 & 0.00 & \bfseries 0.1 & \bfseries 11,268.27 & 0.00 & 0.3 \\
Spiral & 5 & \bfseries 461.53 & 0.00 & 1,758.0 & \bfseries 461.53 & 0.00 & \bfseries 0.7 & \bfseries 461.53 & 0.00 & 1.9 \\
Spiral & 10 & \bfseries 558.25 & 0.00 & 33.0 & \bfseries 558.25 & 0.00 & \bfseries 0.4 & \bfseries 558.25 & 0.00 & 0.7 \\
Tae & 5 & \bfseries 478.85 & 0.00 & 606.0 & 486.98 & 1.70 & \bfseries 0.6 & 486.98 & 1.70 & 2.2 \\
Tae & 15 & \bfseries 684.30 & 0.00 & 391.0 & \bfseries 684.30 & 0.00 & \bfseries 0.7 & \bfseries 684.30 & 0.00 & 1.1 \\
Tae & 20 & \bfseries 711.82 & 0.00 & 32.0 & \bfseries 711.82 & 0.00 & \bfseries 0.2 & \bfseries 711.82 & 0.00 & 0.6 \\
Vehicle & 10 & \bfseries 13,250.13 & 0.00 & 125.0 & \bfseries 13,250.13 & 0.00 & \bfseries 1.4 & \bfseries 13,250.13 & 0.00 & 2.7 \\
Vehicle & 15 & \bfseries 13,334.17 & 0.00 & 28.0 & \bfseries 13,334.17 & 0.00 & \bfseries 0.2 & \bfseries 13,334.17 & 0.00 & 0.3 \\
Wine & 5 & \bfseries 1,284.69 & 0.00 & 165.0 & \bfseries 1,284.69 & 0.00 & \bfseries 0.6 & \bfseries 1,284.69 & 0.00 & 1.5 \\
Wine & 10 & \bfseries 1,285.13 & 0.00 & 95.0 & \bfseries 1,285.13 & 0.00 & \bfseries 0.7 & \bfseries 1,285.13 & 0.00 & 1.5 \\
Wine & 15 & \bfseries 1,285.13 & 0.00 & 83.0 & \bfseries 1,285.13 & 0.00 & \bfseries 1.0 & \bfseries 1,285.13 & 0.00 & 1.7 \\
Wine & 20 & \bfseries 1,290.66 & 0.00 & 40.0 & \bfseries 1,290.66 & 0.00 & \bfseries 0.7 & \bfseries 1,290.66 & 0.00 & 1.1 \\
Zoo & 5 & \bfseries 538.71 & 0.00 & 64.0 & 566.34 & 5.13 & \bfseries 0.7 & 545.27 & 1.22 & 1.8 \\
Zoo & 10 & \bfseries 545.96 & 0.00 & 95.0 & 550.88 & 0.90 & \bfseries 0.8 & 550.88 & 0.90 & 1.6 \\
Zoo & 15 & \bfseries 551.82 & 0.00 & 166.0 & 555.57 & 0.68 & \bfseries 0.7 & 555.57 & 0.68 & 1.9 \\
Zoo & 20 & \bfseries 552.75 & 0.00 & 70.0 & 559.50 & 1.22 & \bfseries 1.0 & 559.50 & 1.22 & 2.2 \\
\midrule Average &  & \bfseries 3,008.42 & 0.00 & 446.2 & 3,012.69 & 0.24 & \bfseries 1.1 & 3,011.69 & 0.12 & 2.7 \\
\bottomrule
\end{tabularx}

	\caption{Comparison of the solutions produced by the PCCC and PCCC-R algorithms to optimal solutions in terms of Inertia (within cluster sum-of-squares). The optimal solutions have been produced by the PC-SOS-SDP Solver. The PCCC and the PCCC-R algorithms find optimal and near-optimal solutions within short running times.}\label{tbl_inertia_experiment1_optimal_solutions}
\end{table*}

\subsection{Comparison to state-of-the-art on well-known benchmark data sets (COL1)}\label{sec_experiments_comparison_col1}
Next, we compare the performance of the PCCC algorithms to the performance of state-of-the-art heuristic algorithms. Also for this comparison, we use the 25 data sets and the corresponding constraint sets from collection COL1. We applied our algorithms (PCCC) and (PCCC-R) and the five algorithms (COPKM, CSC, DILS, LCC, and KMEANS) three times to each instance, every time with a different random seed and a time limit of 1,800 seconds. In Appendix~D of the Online Supplement, we list the settings used to run the algorithms. We recorded for each solution, the ARI value, the Inertia value, the Silhouette coefficient, the number of violated cannot-link constraints, and the elapsed running time. For some instances, the COPKM and the LCC algorithm failed to return a solution (within the time limit). In those cases, we reported an ARI value of 0 and a Silhouette coefficient of -1 and no value for the other metrics.  	

\begin{figure*}
	\centering
	\includegraphics[width=0.5\textwidth]{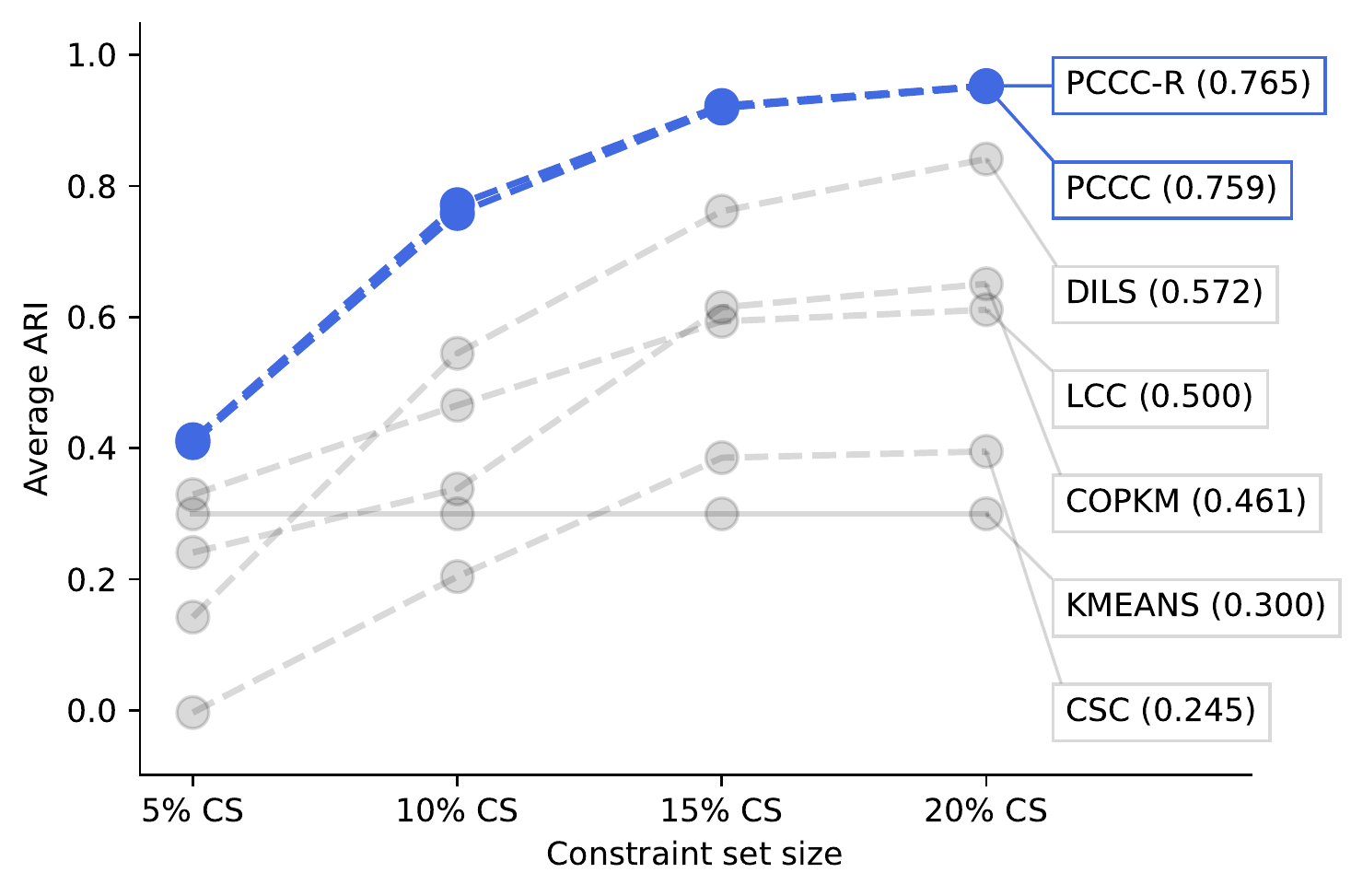}\includegraphics[width=0.5\textwidth]{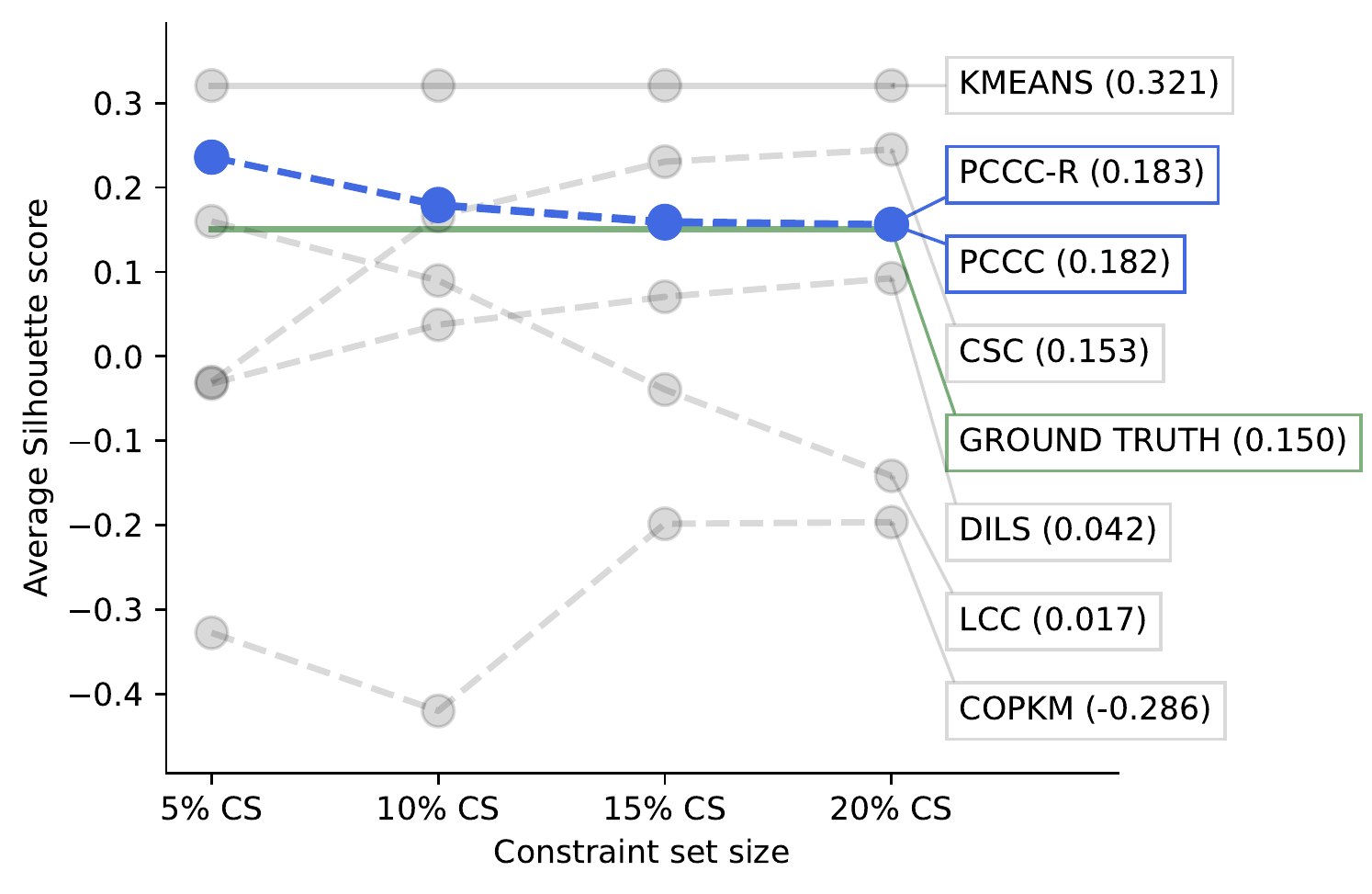}
	\caption{Adjusted Rand Index (left) and Silhouette coefficients (right) for different constraint set sizes, averaged across data sets (from collection COL1) and repetitions. The PCCC-R algorithm and the PCCC algorithm consistently deliver higher ARI values than the state-of-the-art algorithms. The Silhouette coefficients of the PCCC algorithms approach those of the ground truth assignments with increasing size of the constraint sets, demonstrating that the algorithm effectively incorporates the constraints. The values in parentheses behind the label of the algorithm state the average ARI values and Silhouette coefficients across all constraint set sizes. \label{fig_average_ari_silhouette_experiment_1}}
\end{figure*}

Figure~\ref{fig_average_ari_silhouette_experiment_1} shows for each algorithm and each constraint set size, the ARI values (left) and the Silhouette coefficients (right) averaged across data sets and repetitions (random seeds). In terms of ARI values, the PCCC and the PCCC-R algorithm outperform the other algorithms by a substantial margin for all constraint set sizes. The green horizontal line in the plot on the right represents the average Silhouette coefficient of the ground truth. This relatively low value suggests that the clusters in some data sets are not cohesive and well separated. Consequently, this demonstrates that the collection includes datasets that are challenging for clustering algorithms. The PCCC algorithms achieve relatively high Silhouette coefficients when the constraint sets are small. With larger constraint sets, the Silhouette coefficients obtained with the PCCC algorithms approach the Silhouette coefficients of the ground truth assignments. This behavior demonstrates that the PCCC algorithms effectively incorporate the constraints. The state-of-the-art algorithms generally obtain lower Silhouette coefficients except for the CSC algorithm for constraint set sizes 15\% CS and 20\% CS. 

\begin{figure*}
	\centering
	\includegraphics[width=\textwidth]{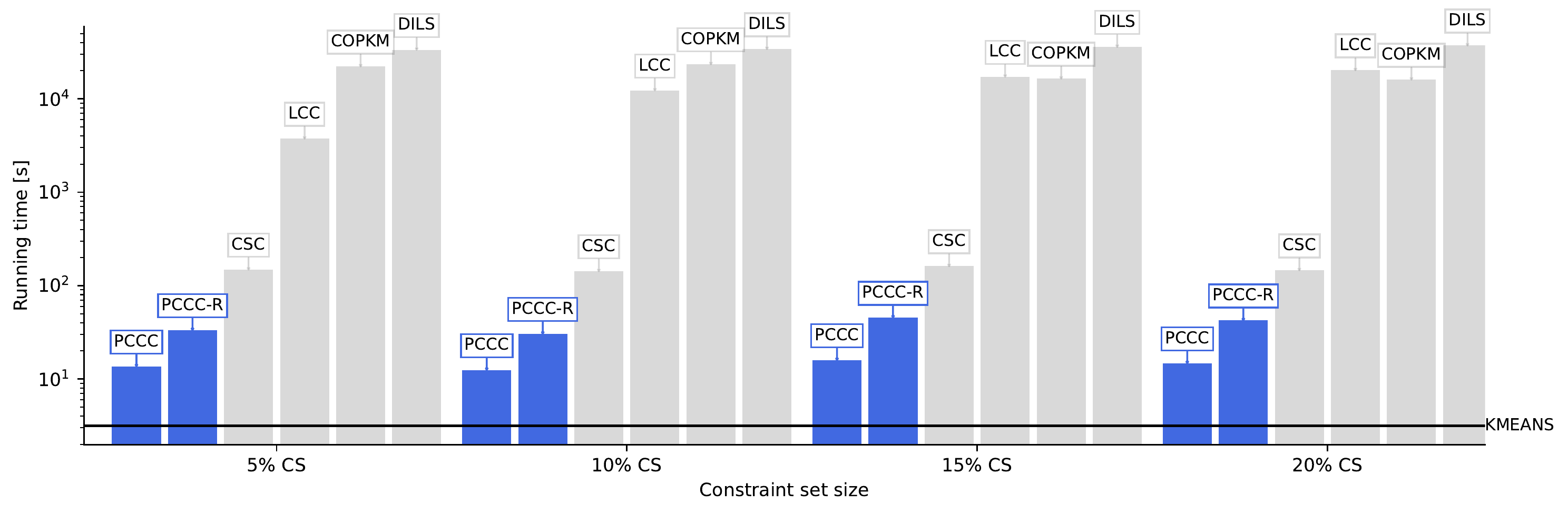}
	\caption{Running time in seconds, on a logarithmic scale, for different constraint set sizes averaged across repetitions and summed up across data sets. The PCCC-R and the PCCC algorithms are considerably faster compared to the state-of-the-art approaches. \label{fig_sum_cpu_experiment_1}}
\end{figure*}

Figure~\ref{fig_sum_cpu_experiment_1} compares the running times of the algorithms for the different constraint set sizes. The running times are averaged across repetitions and summed up across data sets. We replaced nan values with the time limit of 1,800 seconds before aggregating the results. Note that the figure uses a logarithmic scale on the y-axis. The PCCC algorithms are considerably faster than the state-of-the-art approaches. On \href{\githubresults}{\color{blue}GitHub}, we report the detailed ARI values, the Silhouette coefficients, the Inertia values, the number of violated cannot-link constraints, and the running times for all algorithms and data sets in Tables~W5--W24.

\subsection{Impact of complexity parameters on the performance of the PCCC algorithms}\label{sec_experiments_impact_complexity_parameters_col2}
\begin{figure*}
	\centering
	\includegraphics[width=\textwidth]{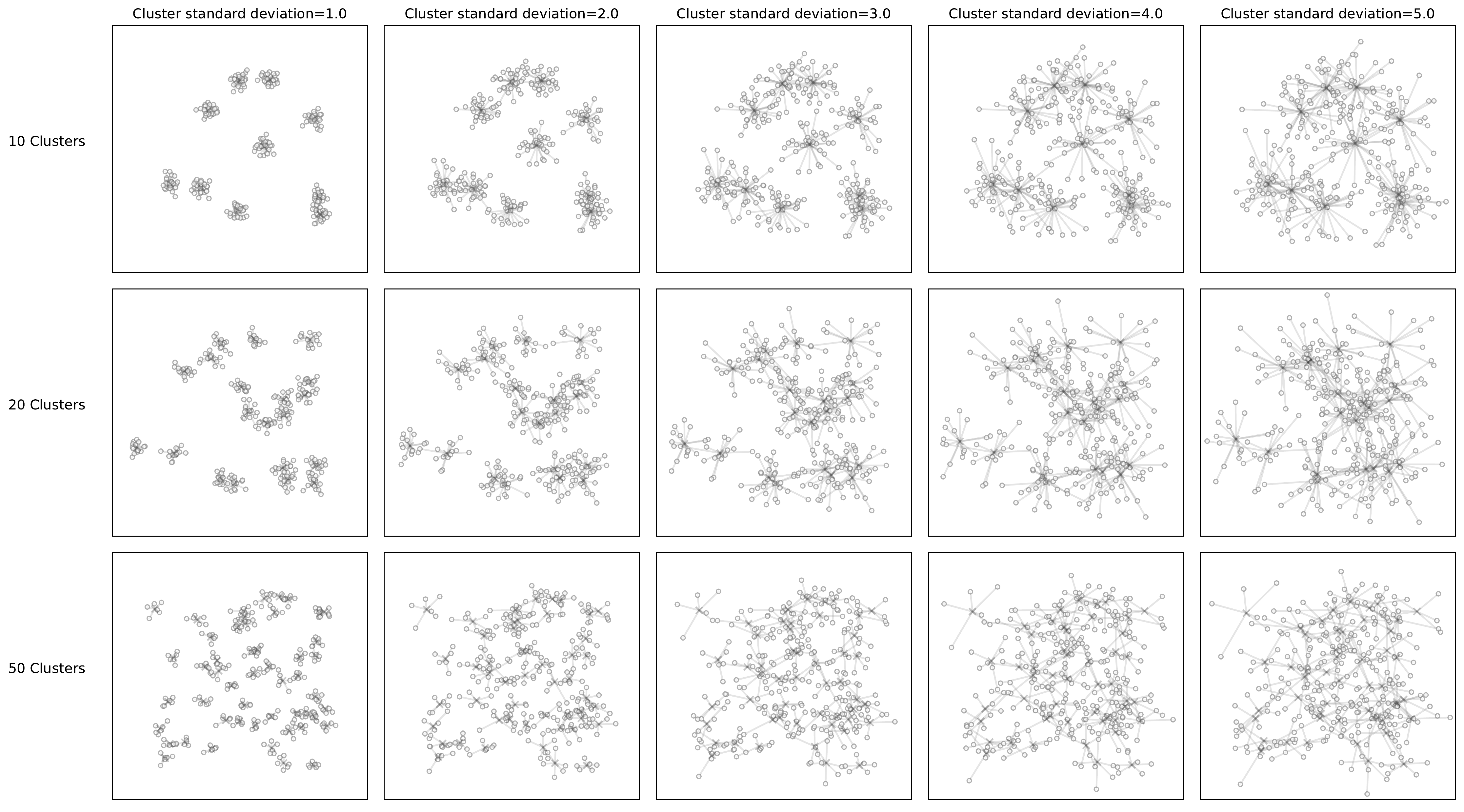}
	\caption{Data sets used to investigate the impact of complexity parameters on the performance of the PCCC algorithms. The gray lines indicate the ground truth assignments and the crosses represent the centers of gravity of the clusters.}\label{fig_impact_of_complexity_parameters_data_sets}
\end{figure*}

In this section, we investigate how the performance of the PCCC algorithms depends on two complexity parameters, namely the chromatic number of the undirected graph $G'$ (see Section~\ref{sec_algorithm_preprocessing}) and the distribution of objects in the data set. We perform this analysis based on the synthetic data sets and corresponding constraints sets from collection COL2. Collection COL2 has 15 data sets which differ with respect to the number of clusters and the standard deviation of the clusters. Figure~\ref{fig_impact_of_complexity_parameters_data_sets} shows the 15 data sets.  

For each data set there are 10 constraint sets of increasing size resulting in 150 problem instances. For this experiment, we only used the cannot-link constraints of the respective constraint set to avoid that the contraction of objects connected by must-link constraints affects the running time required for solving the models. We tested the following versions of the PCCC algorithm:

\begin{itemize}
	\item PCCC: This is the baseline version that we used in the previous section (see Appendix~D of the Online Supplement for details). Here we increased the solver time limit to 3,600 seconds to analyze the impact of the complexity parameters on the time required to solve the full model in the assignment step.  
	\item PCCC-N2-S: In this version, we provide the cannot-link constraints as soft constraints with $w_{ij}=1$ for $\{i, j\} \in E^\textup{SCL}$. The model-size reduction technique is applied with $q=2$. Parameter $P$ is computed internally with the default procedure. Model (R($q$)MBLP) is solved with a solver time limit of 3,600 seconds. The initial positions of the cluster centers are determined with the k-means++ algorithm of \cite{arthur2006k}. The algorithm uses the Gurobi Optimizer 11.0.2 as solver. 
\end{itemize}

\begin{figure*}
	\centering
	\includegraphics[width=\textwidth]{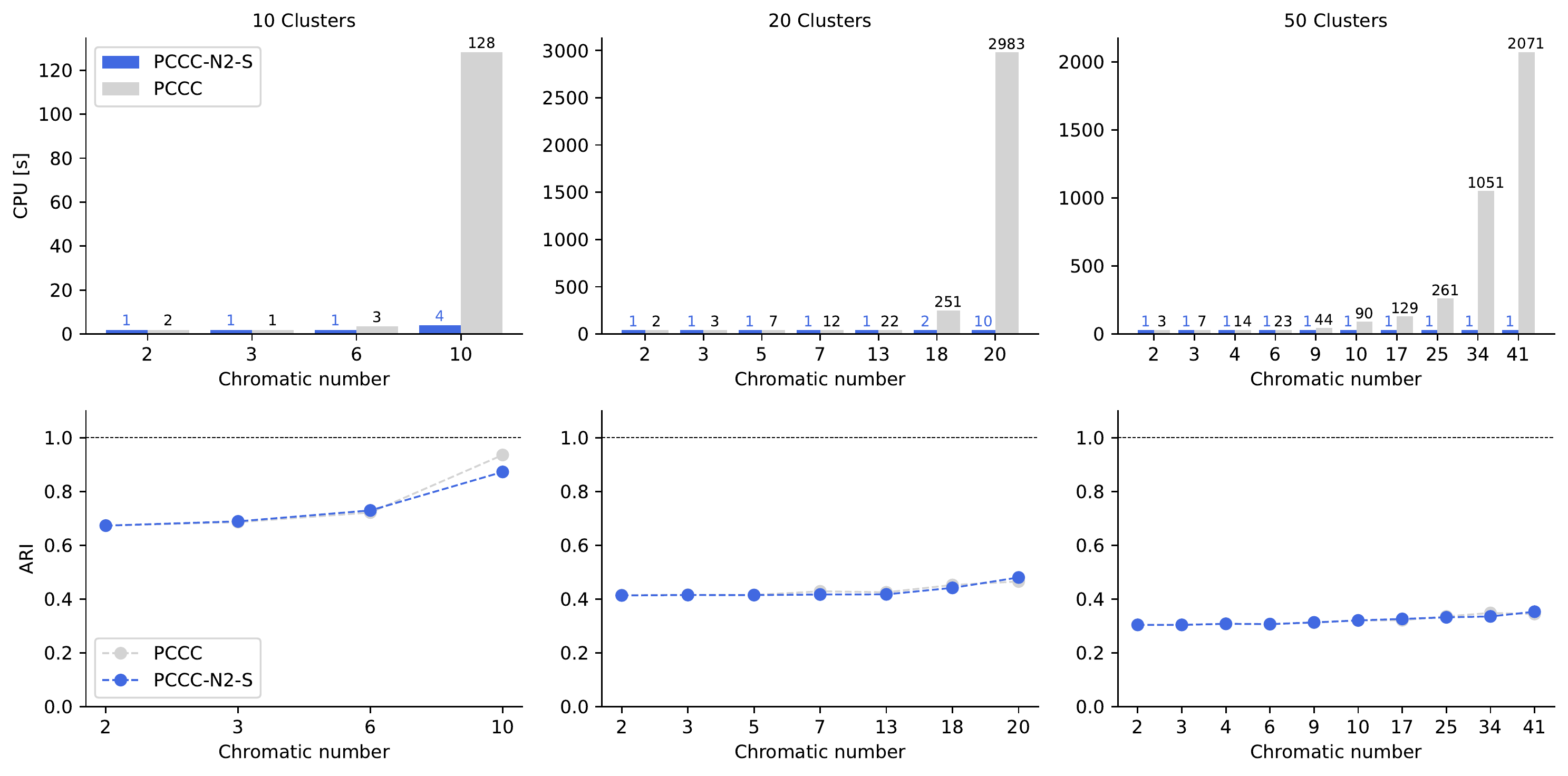}
	\caption{Impact of number of clusters and chromatic numbers on running time and ARI values. While the running time of the PCCC algorithm increases with increasing chromatic number, the running time of the PCCC-N2-S algorithm is consistently low and does not seem to be affected by this complexity parameter. Please note that we set a minimum bar height in the bar plots to ensure that all bars are visible. \label{fig_impact_of_chromatic_number}}
\end{figure*}

We applied the PCCC algorithm and the PCCC-N2-S algorithm three times to each instance, each time with a different random seed and a time limit of 3,600 seconds. Note that the instances that differ only with respect to the distribution of the objects have the same graph $G'$. To each unique graph $G'$, we applied the CP-SAT solver from Google OR-Tools (see~\citealt{cpsatlp}) to determine its chromatic number. We imposed a solver time limit of 7,200 seconds for each graph. For 23 graphs, the chromatic number could be computed within the time limit. For the other 7 graphs, we obtained upper bounds on the chromatic number that are in our opinion sufficiently tight (max absolute gap is 4 and the max relative gap is 13.6\%) to study the impact of the chromatic number on the performance of the PCCC algorithms. Figure~\ref{fig_impact_of_chromatic_number} shows for different numbers of clusters, the relationship between the chromatic number and the running time (plots in first row) and the relationship between the chromatic number and the ARI values (plots in second row). Some graphs had the same chromatic number for different constraint sets. We list on the horizontal axes the unique values for the chromatic numbers obtained and report the respective average running time or average ARI values (average across constraint sets and data sets with different cluster standard deviation). The running time of the PCCC algorithm increases with increasing chromatic number of the graph. Interestingly, the running time is particularly high when the chromatic number matches the number of clusters. Additionally, the running time is influenced more by how close the chromatic number is to the number of clusters than by the absolute value of the chromatic number. The running time of the PCCC-N2-S algorithm does not seem to depend on the chromatic number and is consistently low. The detailed results on \href{\githubresults}{\color{blue}GitHub} reveal that the impact of the chromatic number on the running time is weaker when the clusters are well separated. This is because in the case of well-separated clusters, assignments with low Inertia tend to satisfy the cannot-link constraints.  

\begin{figure*}
	\centering
	\includegraphics[width=\textwidth]{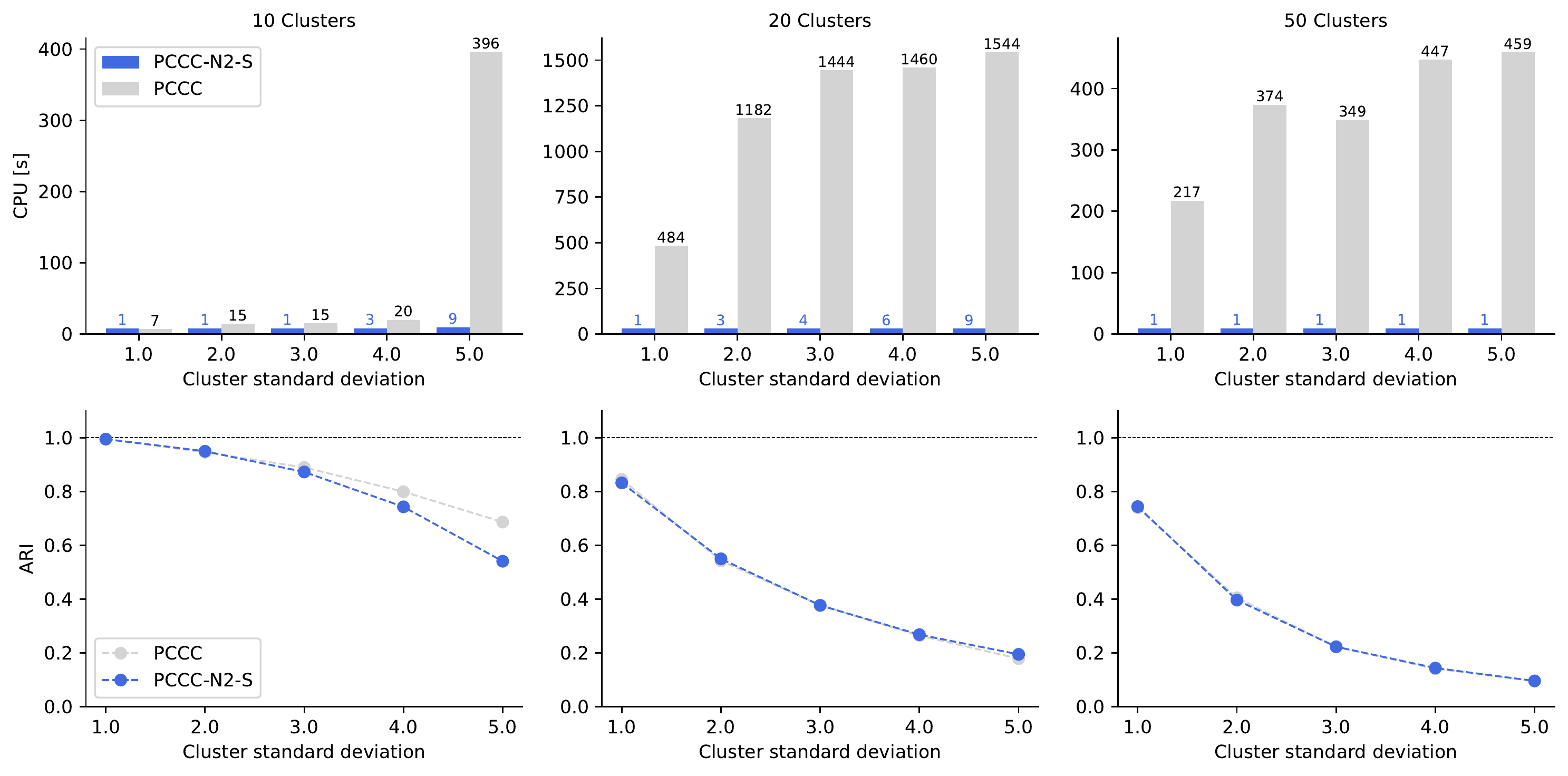}
	\caption{Impact of number of clusters and cluster standard deviation on running time and ARI values. While the running time of the PCCC algorithm increases with increasing cluster standard deviation, the running time of the PCCC-N2-S algorithm is consistently low and much less affected by this complexity parameter. The ARI values obtained with both algorithms strongly depend on the distribution of the objects. Please note that we set a minimum bar height in the bar plots to ensure that all bars are visible. \label{fig_impact_of_cluster_standard_deviation}}
\end{figure*}

Next, we analyze how the performance of the PCCC algorithms depends on the distribution of objects. Notably, the problem is known to become more challenging (even without constraints) when the objects cannot be clustered into well-separated groups. Figure~\ref{fig_impact_of_cluster_standard_deviation} shows for different numbers of clusters, the relationship between the cluster standard deviation and the running time (plots in first row) and the relationship between the cluster standard deviation and the ARI values (plots in the second row). We list on the horizontal axes the different cluster standard deviations and report the average running time and the average ARI values (average across all constraint sets). While the running time of the PCCC algorithm increases with increasing cluster standard deviation, the running time of the PCCC-N2-S algorithm is consistently low and does not seem to be affected by this complexity parameter. The ARI values obtained with both algorithms strongly depend on the distribution of the objects. On \href{\githubresults}{\color{blue}GitHub}, we report the detailed ARI values, the Silhouette coefficients, the Inertia values, the number of violated cannot-link constraints, and the running times for all algorithms and data sets in Tables~W25--W74.

\subsection{Comparison to state-of-the-art approaches on synthetic data sets (COL3)}\label{sec_experiments_comparison_col3}
The synthetic data sets of collection COL3 allow us to investigate how the performance of the PCCC and the benchmark algorithms depends on the size of the data sets in terms of the number of objects and the number of clusters. We use the same state-of-the-art algorithms and the same experimental design as in Section~\ref{sec_experiments_comparison_col1}, except that we increased the time limit to 3,600 seconds. If an algorithm failed to return a solution (within the time limit), we reported an ARI value of 0. We noticed that the LCC algorithm stops with a runtime error when the constraint set is empty. This is why the LCC algorithm did not return any solutions for the constraint sets of size 0\% CS. We tested the following versions of the PCCC algorithm:

\begin{itemize}
	\item PCCC: This is the baseline version that we used in the previous comparison (see Appendix~D of the Online Supplement). Here we set the solver time limit to 30 seconds. 
	\item PCCC-N2-S: In this version, we provide the must-link constraints as hard constraints and the cannot-link constraints as soft constraints with $w_{ij}=1$ for $\{i, j\} \in E^\textup{SCL}$. The model-size reduction technique is applied with $q=2$. Parameter $P$ is computed internally with the default procedure. Model (R($q$)MBLP) is solved with a solver time limit of 30 seconds. The initial positions of the cluster centers are determined with the k-means++ algorithm of \cite{arthur2006k}. The algorithm uses the Gurobi Optimizer 11.0.2 as solver. 
	\item PCCC-N5-S: Corresponds to PCCC-N2-S, but with $q=5$. \textbf{Comment:} We tested additional versions (e.g., PCCC-N3-S, PCCC-N4-S), but their performance was similar to the performance of either PCCC-N2-S or PCCC-N5-S.
	\item PCCC-N2-S-RD: Corresponds to PCCC-N2-S, but with cluster repositioning and dynamic enlargement of search space with $\gamma=500$ and $\delta=10$.	
\end{itemize}

\begin{figure*}
	\centering
	\includegraphics[width=0.5\textwidth]{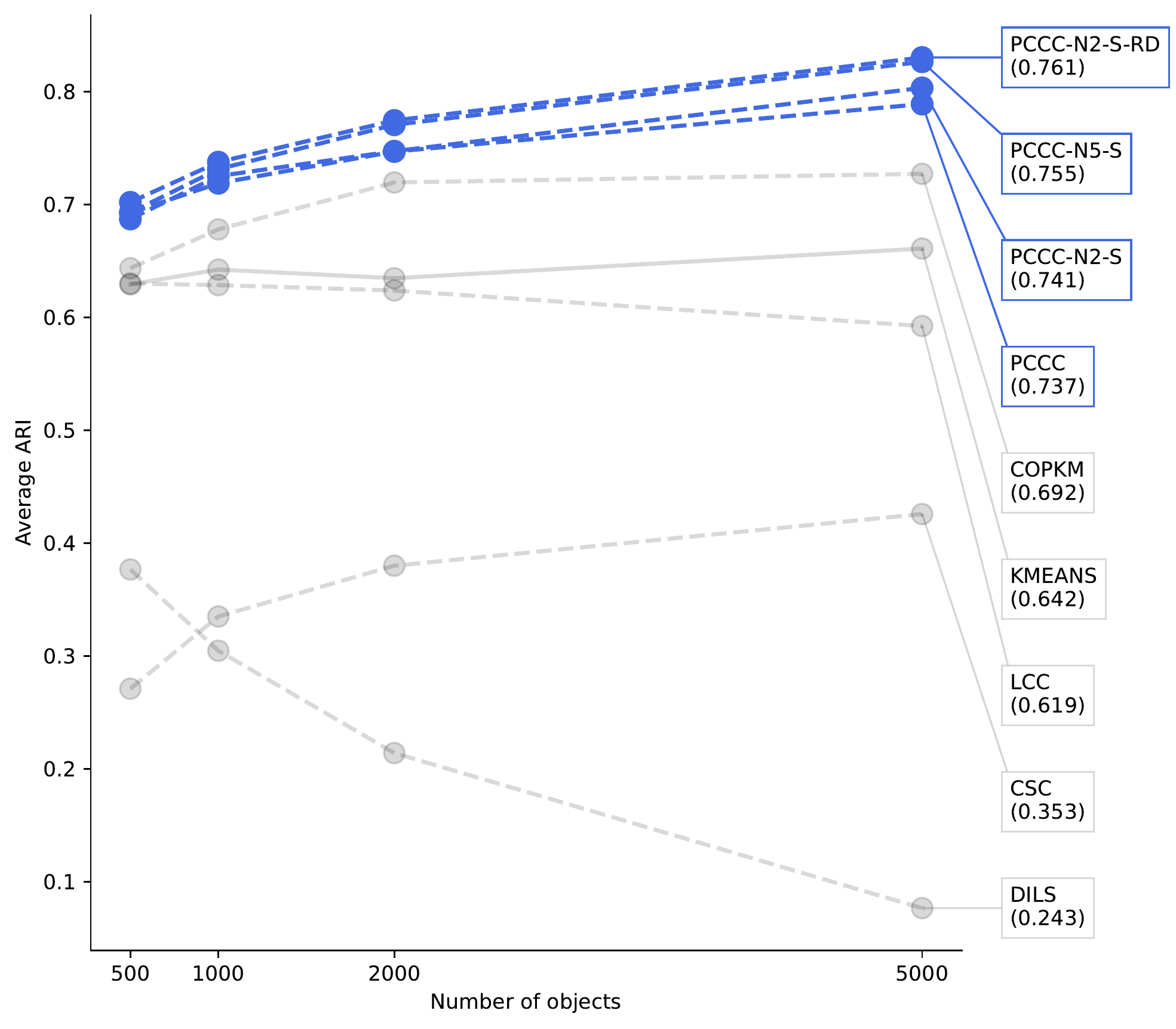}\includegraphics[width=0.5\textwidth]{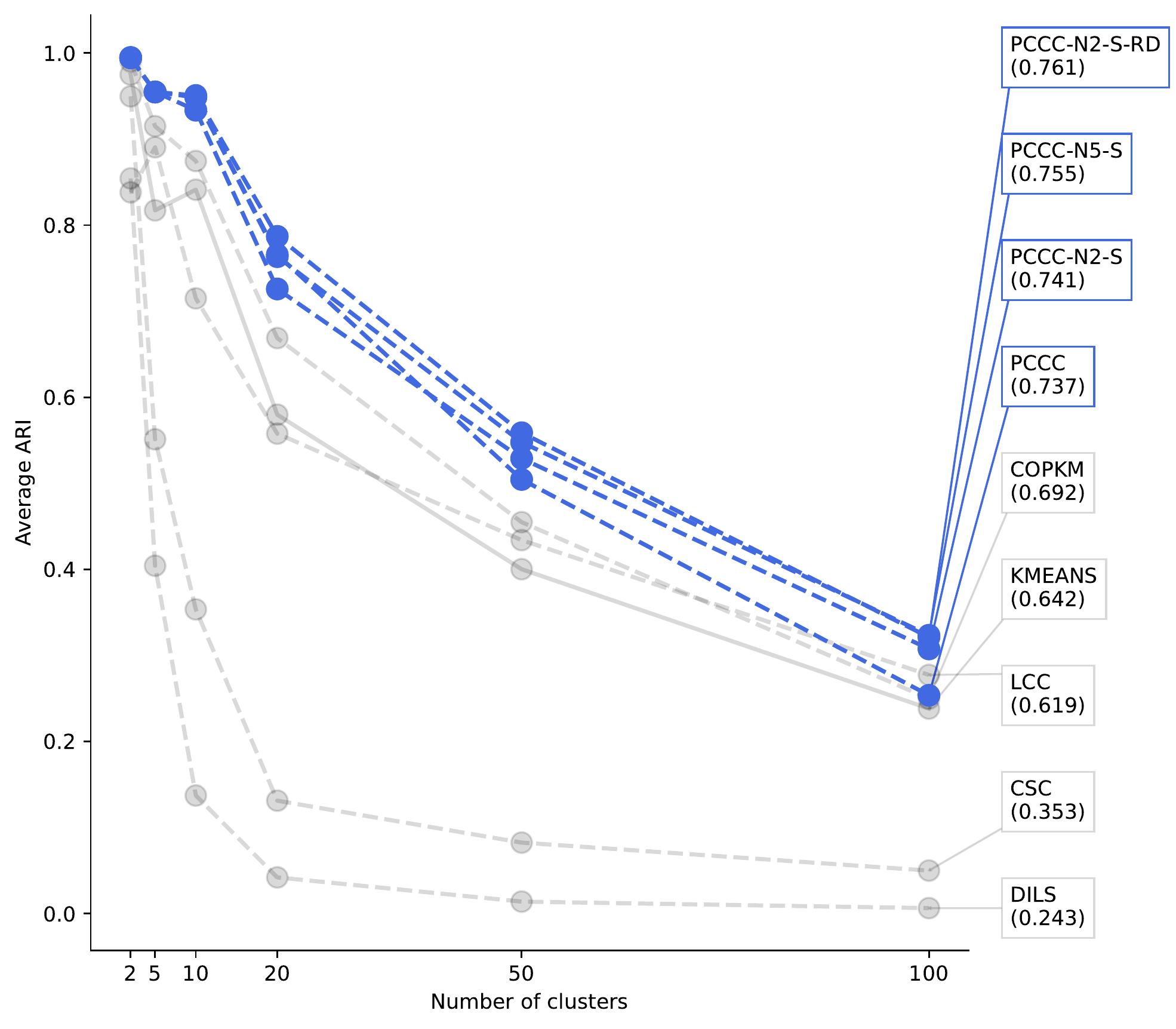}
	\caption{Average Adjusted Rand Index for increasing numbers of objects (left) and increasing number of clusters (right). The PCCC versions deliver leading performance across all instance sizes. The values in parentheses below the label of the algorithm state the average ARI values across all data sets. \label{fig_average_ari_experiment_2}}
\end{figure*}

Figure~\ref{fig_average_ari_experiment_2} shows the average ARI values that were obtained for data sets with different numbers of objects (left) and different numbers of clusters (right). For each data set size, we report the ARI value averaged across data sets, constraint sets, and repetitions (random seeds). We can see that the versions of the PCCC algorithm perform best across all data set sizes. The versions of the PCCC algorithm that use the model-size reduction technique and treat the cannot-link constraints as soft constraints (PCCC-N2-S, PCCC-N5-S, and PCCC-N2-S-RD) are particularly effective when the number of clusters is large. For these instances, the full model becomes too large and Gurobi no longer finds high quality assignments within the given time limit which explains why the PCCC algorithms that use the model-size reduction technique (PCCC-N2-S, PCCC-N5-S, and PCCC-N2-S-RD) achieve higher ARI values than the PCCC algorithm. The low average ARI values of the DILS algorithm in this experiment are surprising. It turns out that the performance of the DILS algorithm decreases rapidly when the number of objects or clusters increases. Also the ARI values of the CSC algorithm decrease considerably when the number of clusters increases. 

\begin{figure*}
	\centering
	\includegraphics[width=\textwidth]{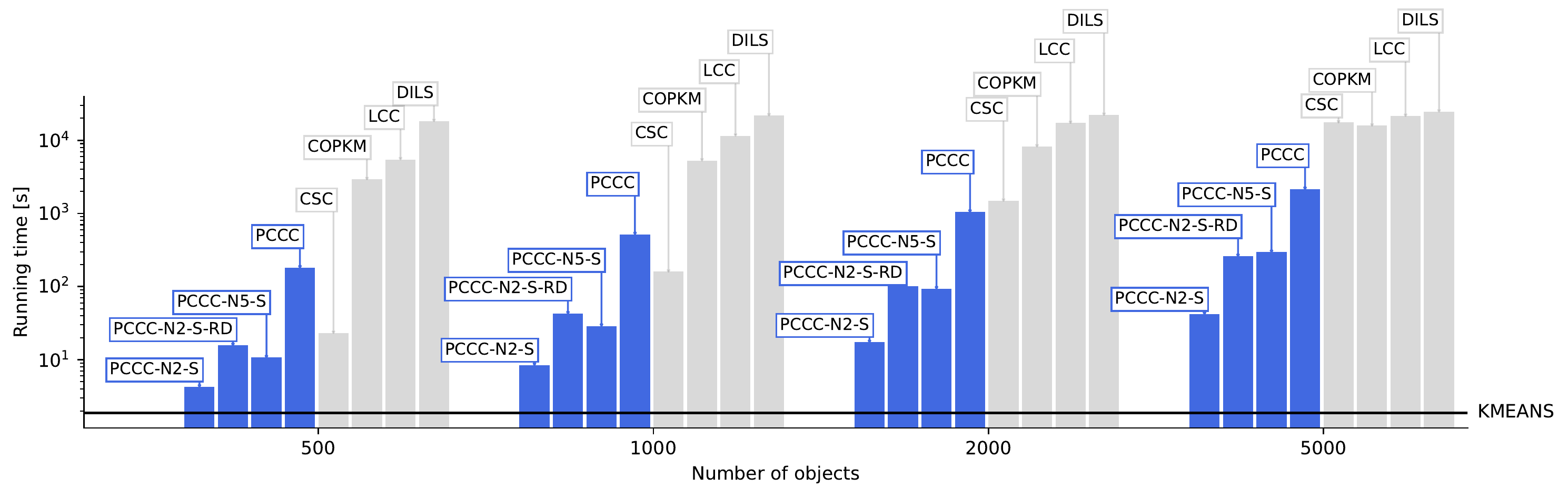}
	\caption{Running time in seconds, on a logarithmic scale, for different numbers of objects averaged across repetitions and constraint sets, and summed up across data sets (from collection COL3). The PCCC versions (in particular those that treat the cannot-link constraints as soft constraints) are much faster than the state-of-the-art approaches. \label{fig_sum_cpu_experiment_2_n_objects}}
\end{figure*}

\begin{figure*}
	\centering
	\includegraphics[width=\textwidth]{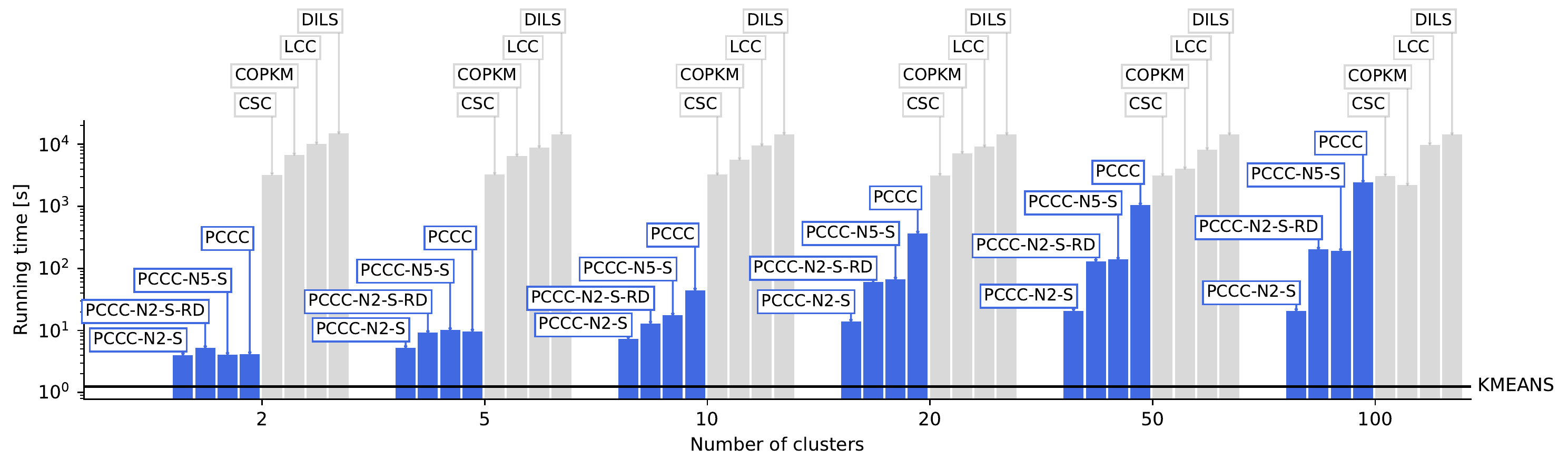}
	\caption{Running time in seconds, on a logarithmic scale, for different numbers of clusters averaged across repetitions and constraint sets, and summed up across data sets (from collection COL3). The PCCC versions (in particular those that treat the cannot-link constraints as soft constraints) are much faster than the state-of-the-art approaches. \label{fig_sum_cpu_experiment_2_n_clusters}}
\end{figure*}

Figures~\ref{fig_sum_cpu_experiment_2_n_objects} and \ref{fig_sum_cpu_experiment_2_n_clusters} compare the running time of the tested algorithms on data sets with different numbers of objects and different numbers of clusters, respectively. Both figures report the running time averaged across repetitions and constraint sets, and summed up across data sets. We replaced nan values with the time limit of 3,600 seconds before aggregating the results. The versions of the PCCC algorithm that use the model-size reduction technique and treat the cannot-link constraints as soft constraints are orders of magnitudes faster than the state-of-the-art approaches and thus scale much better to instances with large number of objects and clusters. Also, it appears that the trade-off between speed and quality can be effectively controlled with parameter $q$. Version PCCC-N2-S with $q=2$ is faster but delivers slightly lower ARI values compared to version PCCC-N5-S with $q=5$. In Appendix~E of the Online Supplement, we investigate in detail for which type of instances the model-size reduction technique is particularly effective. We can conclude that when the number of clusters is large (50 or more), the model-size reduction technique becomes highly effective. On \href{\githubresults}{\color{blue}GitHub}, we report the detailed ARI values, the Silhouette coefficients, the Inertia values, the number of violated cannot-link constraints, and the running times for all algorithms and data sets in Tables~W75--W99.

\subsection{Comparison to state-of-the-art approaches on large-sized data sets (COL4)}\label{sec_experiments_comparison_col4}
To further assess the scalability of the PCCC-N2-S, the PCCC-N5-S, and the PCCC-N2-S-RD versions, we conducted another experiment with the large-sized data sets from collection COL4 that comprise up to 70,000 objects, 3,072 features, and 100 clusters. We applied all state-of-the-art algorithms and the three PCCC versions to these data sets. Table~\ref{tbl_ari_experiment3} reports for each data set and constraint set size 5\% CS the ARI values of the algorithms averaged across repetitions. A hyphen ''--" means that the algorithm failed to return a solution within the time limit of 3,600 seconds. The PCCC versions produce solutions with considerably higher ARI values compared to the solutions obtained by the standard KMEANS algorithm while the state-of-the-art algorithms mostly fail to return a solution within the time limit. These results demonstrate the superiority of the proposed algorithms in terms of scalability. The benefit from using the cluster repositioning and the dynamic enlargement of the search space is reflected in the performance of version PCCC-N2-S-RD which clearly outperforms both other versions of the PCCC algorithm. The detailed results for the individual constraint set sizes are reported on \href{\githubresults}{\color{blue}GitHub} in the Tables~W100--W119. 

\begin{table*}
	\centering
	\scriptsize
	\renewcommand{\tabcolsep}{3pt}
	\begin{tabularx}{\textwidth}{Xrrrrrrrrrrr}
\toprule
{} & & & & \multicolumn{7}{c}{5\% CS} &     $$ \\
\cmidrule(lr){5-11} &  Objects & Features & Clusters &       PCCC-N2-S &       PCCC-N5-S &    PCCC-N2-S-RD &           COPKM &             LCC &             CSC &            DILS & KMEANS \\
Dataset       &          &          &          &                 &                 &                 &                 &                 &                 &                 &        \\
\midrule
Banana        &    5,300 &        2 &        2 &  \textbf{1.000} &  \textbf{1.000} &  \textbf{1.000} &             -- &             -- &           0.996 &           0.100 &  0.017 \\
Letter        &   20,000 &       16 &       26 &           0.587 &           0.656 &  \textbf{0.730} &             -- &             -- &             -- &           0.000 &  0.149 \\
Shuttle       &   57,999 &        9 &        7 &           0.998 &  \textbf{1.000} &  \textbf{1.000} &             -- &             -- &             -- &             -- &  0.411 \\
CIFAR 10      &   60,000 &    3,072 &       10 &           0.525 &           0.672 &  \textbf{1.000} &             -- &             -- &             -- &             -- &  0.040 \\
CIFAR 100     &   60,000 &    3,072 &      100 &           0.262 &           0.227 &  \textbf{0.326} &             -- &             -- &             -- &             -- &  0.021 \\
MNIST         &   70,000 &      784 &       10 &           0.744 &           0.749 &  \textbf{1.000} &             -- &             -- &             -- &             -- &  0.312 \\
\midrule Mean &   &   &   &           0.686 &           0.717 &  \textbf{0.843} &  0.000$^{\ast}$ &  0.000$^{\ast}$ &  0.166$^{\ast}$ &  0.017$^{\ast}$ &  0.159 \\
\bottomrule
\end{tabularx}

\footnotesize $^{\ast}$Nan values (--) are replaced with 0 before computing the mean.
	\caption{Average ARI values for large-sized data sets and constraint set size 5\% CS. A hyphen ''--" means that the algorithm failed to return a solution within the time limit of 3,600 seconds.}\label{tbl_ari_experiment3}
\end{table*}

\section{Constrained clustering with confidence values}\label{sec_experiments_comparison_noisy}
Finally, we conduct an experiment to analyze the benefits of accounting for confidence values. To this end, we use the three small data sets Appendicitis, Moons, and Zoo, from collection COL1. With small data sets, the assignment problems of the PCCC versions can be solved optimally within seconds. Hence, the reported results do not change if we increase the solver time limit. We generated 24 noisy constraint sets for each of the four data sets Appendicitis, Moons, and Zoo, resulting in 72 problem instances. These constraint sets may contain incorrect constraints that do not agree with the ground truth labels. Table~14 in Appendix~F of the Online Supplement provides information on the number of constraints that do not agree with the ground truth labels for each constraint set. The noisy constraints are associated with confidence values that indicate their reliability. A detailed description of how we generated the noisy constraint sets can be found in Appendix~F of the Online Supplement. From the state-of-the-art algorithms, only the CSC algorithm can take as input confidence values for the constraints. We applied the following algorithms three times to each problem instance: 

\begin{itemize}
	\item PCCC: Corresponds to the baseline version that we used in the previous comparisons (see Appendix~D of the Online Supplement). 
	\item PCCC-S: Corresponds to PCCC, but all pairwise constraints are provided as soft constraints with $w_{ij}=1$ and parameter $P$ is specified as the average distance between a node and a cluster center.
	\item PCCC-W: Corresponds to PCCC, but all pairwise constraints are provided as soft constraints and each constraint receives the confidence value $w_{ij}$ that was computed by the generation procedure described in Appendix~F of the Online Supplement. Parameter $P$ is specified as the average distance between a node and a cluster center.
	\item CSC-W: Corresponds to the CSC algorithm, but in this version of the algorithm, the constraint matrix $Q$ contains entry $w_{ij}$ for every noisy must-link constraint and entry $-w_{ij}$ for every noisy cannot-link constraint. All other settings are chosen as in version CSC.
	\item KMEANS: We set parameter n\_init (number of initializations) to 1 and otherwise used the default settings.	
\end{itemize} 

\begin{figure*}
	\centering
	\includegraphics[width=0.9\textwidth]{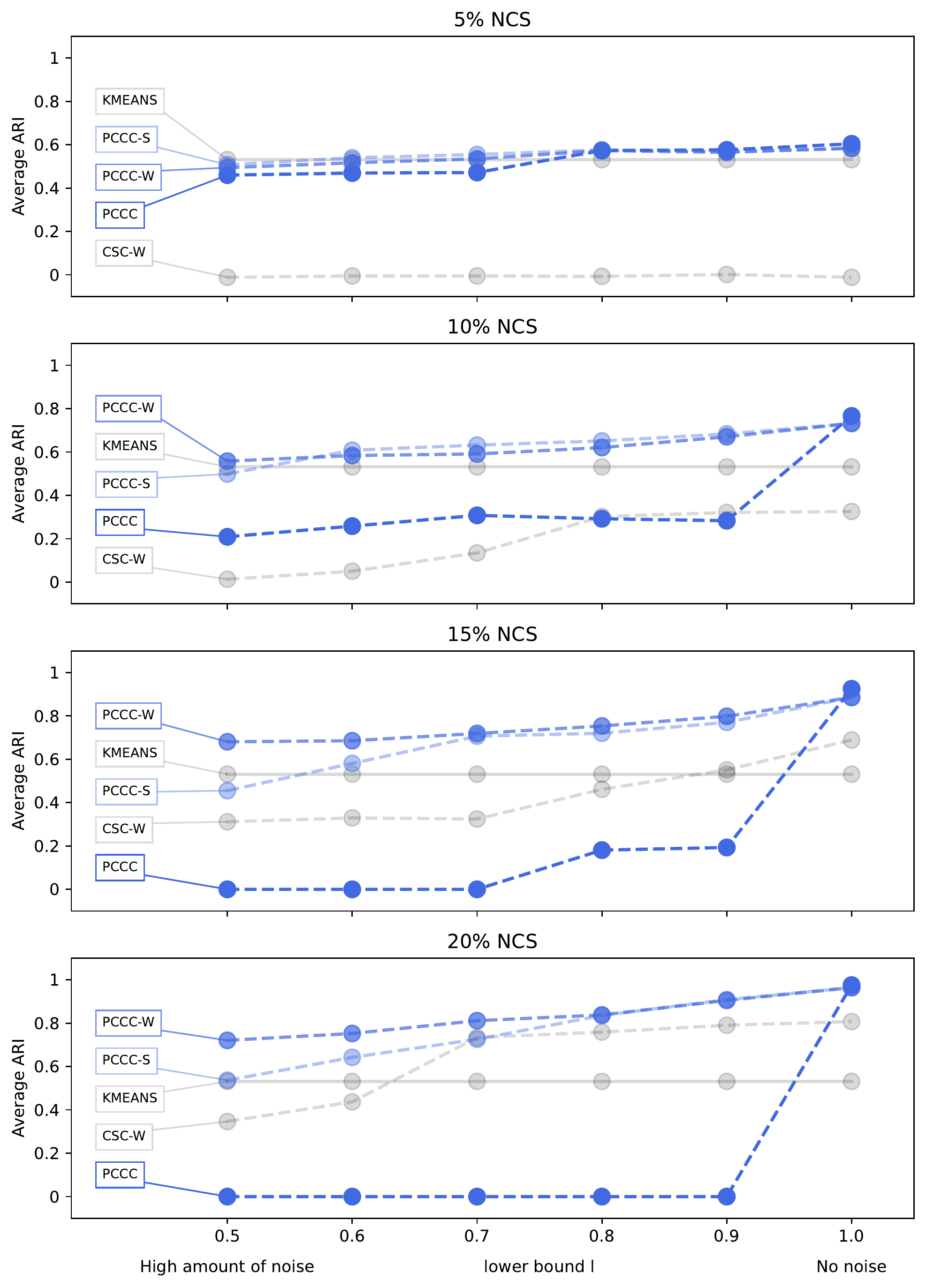}
	\caption{Adjusted Rand Index for different noisy constraint sets, averaged across data sets and repetitions. Overall, the confidence-based variant (PCCC-W) delivers the highest ARI values, especially when the constraint sets are large and contain a high level of noise (amount of noise decreases from left to right).}\label{fig_ari_average_experiment_4}
\end{figure*}

Figure~\ref{fig_ari_average_experiment_4} shows the ARI values averaged across data sets and repetitions for the different constraint sets. The figure contains a separate line plot for each of the four constraint set sizes, and the horizontal axes of the plots represent the amount of noise in the constraint sets (amount of noise decreases from left to right). Specifically, the horizontal axis represents the lower bound $l$ that was provided as an input parameter to the constraint set generation procedure. On \href{\githubresults}{\color{blue}GitHub}, Tables W120--W123 provide the detailed numerical results. When the noisy constraint sets are small (5\% CS), the KMEANS and the PCCC versions deliver considerably better results than the state-of-the-art approach. For medium-sized constraint sets (10\% CS), the soft version (PCCC-S) and the confidence-based version (PCCC-W) perform similarly well. Both versions outperform the version that treats the noisy constraints as hard constraints (PCCC) and the state-of-the-art approach (CSC-W), which both have substantially lower average ARI values compared to the KMEANS algorithm. For large constraint sets (15\% CS and 20\% CS), the confidence-based variant (PCCC-W) consistently delivers the highest average ARI values compared to all other algorithms. The outperformance is substantial when the amount of noise is large, i.e., when the lower bound $l$ is low. These results demonstrate that accounting for confidence values is highly beneficial when the constraint sets are large and noisy. Another insight from Figure~\ref{fig_ari_average_experiment_4} is that considering noisy constraints as hard constraints negatively impacts performance.

\section{Conclusions}\label{sec_conclusions}
We introduce the PCCC algorithm for semi-supervised clustering. Our algorithm is center-based and iterates between an object assignment and a cluster center update step. The key idea is to use integer programming in the object assignment step to accommodate additional information in the form of constraints. Using integer programming makes the algorithm more flexible than state-of-the-art algorithms. For example, the PCCC algorithm can accommodate the additional information in the form of hard or soft constraints, depending on the clustering application. Moreover, the user can specify a confidence value for each soft constraint that reflects the degree of belief in the corresponding information. We demonstrate that this flexibility is beneficial when the additional information is noisy. Key feature of the approach are a cluster repositioning procedure and a model-size reduction technique based on kd-trees. These techniques enable the algorithm to quickly generate high-quality solutions for instances with up to 60,000 objects, 3,072 features, 100 clusters, and hundreds of thousands of cannot-link constraints. In a comprehensive computational analysis, we demonstrate the superiority of the PCCC algorithm over the state-of-the-art approaches for semi-supervised clustering. These findings align with recent literature highlighting the significant role of optimization techniques from the field of Operations Research in various important machine learning settings (see \citealt{gambella2021optimization}). For future research, we plan to apply the proposed framework to related constrained clustering problems where the additional information is given in the form of cardinality, balance, or fairness constraints. We also plan to develop a version of the PCCC algorithm that can be applied to the discrete $k$-median clustering problem where the cluster centers are required to be chosen among the objects (see \citealt{seref2014mathematical}). Another promising research direction is to extend the scope of the PCCC algorithm to non-spherical clusters by combining it with kernel approximation techniques such as the one proposed in \cite{wang2019scalable}. 

\section*{Acknowledgement}
The second author's research was supported by the AI Institute NSF Award 2112533.	
\bibliography{literature.bib}

\clearpage

\end{document}